\pdfoutput=1
\documentclass[11pt]{article}
\usepackage[]{acl}
\usepackage{times}
\usepackage{latexsym}
\usepackage{booktabs}
\usepackage{hyperref}
\usepackage[T1]{fontenc}
\usepackage[utf8]{inputenc}
\usepackage{microtype}
\usepackage[normalem]{ulem}

\usepackage{fancyvrb}
\usepackage{soul}
\usepackage{url}
\usepackage[hidelinks]{}
\usepackage[utf8]{inputenc}
\usepackage{caption}
\usepackage{graphicx}
\usepackage{amsmath}
\usepackage{booktabs}
\usepackage{tikz}
\tikzset{every picture/.style={remember picture}}
\usetikzlibrary{tikzmark}
\usepackage{multirow}
\usepackage{makecell}
\usepackage{hhline}
\usepackage{paralist}
\usepackage{enumitem}

\newcommand\sect[1]{\S\ref{#1}}

\title{Reframing Human-AI Collaboration for \\ Generating Free-Text Explanations}

\author{Sarah Wiegreffe\textsuperscript{$\clubsuit$} \hspace*{10mm} Jack Hessel\textsuperscript{$\dagger$} \hspace*{10mm} Swabha Swayamdipta\textsuperscript{$\dagger$} \\
\textbf{Mark Riedl\textsuperscript{$\clubsuit$} \hspace*{10mm} Yejin Choi\textsuperscript{$\Diamond\dagger$}} \\
  \textsuperscript{$\clubsuit$}School of Interactive Computing, Georgia Institute of Technology \\
    \textsuperscript{$\dagger$}Allen Institute for Artificial Intelligence\\
  \textsuperscript{$\Diamond$}Paul G. Allen School of Computer Science and Engineering, University of Washington\\
  \texttt{saw@gatech.edu, riedl@cc.gatech.edu, \{jackh,swabhas,yejinc\}@allenai.org} \\}

\begin{document}
\maketitle
\begin{abstract}
Large language models are increasingly capable of generating fluent-appearing text with relatively little task-specific supervision.
But can these models accurately explain classification decisions?
We consider the task of generating free-text explanations using human-written examples in a few-shot manner.
We find that (1) authoring higher quality prompts results in higher quality generations; and (2) surprisingly, in a head-to-head comparison, crowdworkers often prefer explanations generated by GPT-3 to crowdsourced explanations in existing datasets.
Our human studies also show, however, that while models often produce factual, grammatical, and sufficient explanations, they have room to improve along axes such as providing novel information and supporting the label.
We create a pipeline that combines GPT-3 with a supervised filter that incorporates binary acceptability judgments from humans in the loop. 
Despite the intrinsic subjectivity of acceptability judgments, we demonstrate that acceptability is partially correlated with various fine-grained attributes of explanations.
Our approach is able to consistently filter GPT-3-generated explanations deemed acceptable by humans.
\end{abstract}
\section{Introduction}
\label{sec:intro}

As natural language understanding tasks have become increasingly complex, the field of explainable NLP has embraced explanations written in free-form natural language.
In contrast to extractive explanations that highlight tokens in the input, free-text explanations provide a natural interface between machine computation and human end-users \cite{hendricks2016generating, camburu2018snli}.
The dominant paradigm for producing free-text explanations is via direct supervision, i.e., training an autoregressive, generative language model to predict human-authored explanations directly 
\begin{figure}[ht!]
\centering
  \includegraphics[width=0.90\columnwidth]{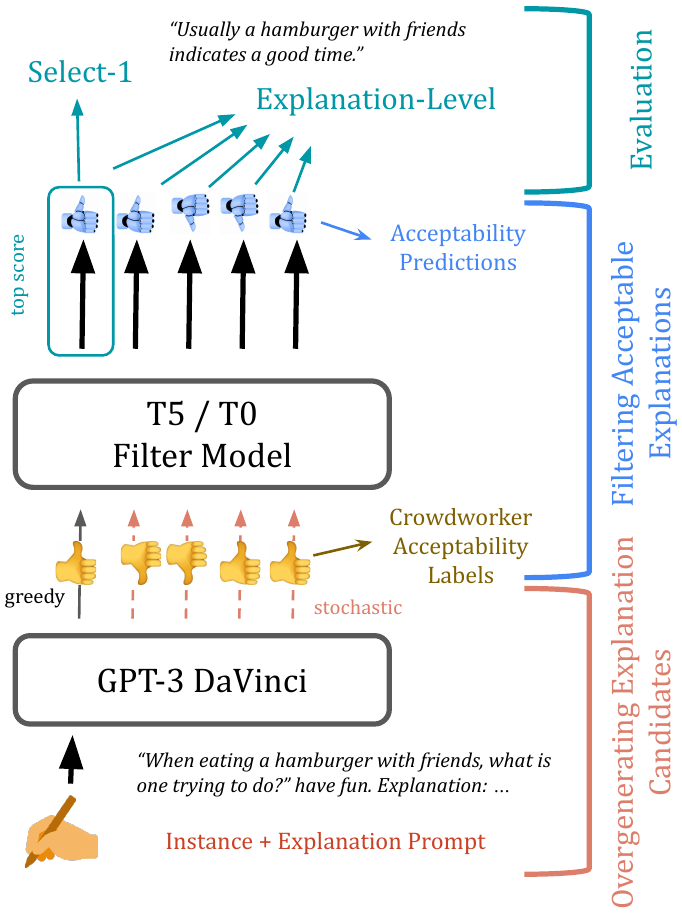}
\caption{
Illustration of our overgeneration + filtration pipeline for producing human acceptable explanations for CommonsenseQA and SNLI (see examples in \autoref{tab:examples}).
Authors of this work write explanations to prompt GPT-3, generating 5 explanations per instance.
An acceptability filter, trained on human binary acceptability judgments, determines which of these generated explanations are plausible.
Evaluation is performed at both the explanation and the instance level.
}
\label{fig:filter_model}
\end{figure}
\cite[\emph{i.a.}]{kim2018textual, park2018multimodal,  ehsan2018rationalization, narang2020wt5, wiegreffe-etal-2021-measuring}.

However, collecting high-quality written explanations to serve as supervision is difficult and expensive.
More than $70\%$ of existing free-text explanation datasets are crowdsourced \cite{wiegreffe2021teach}, and even the most meticulous crowdsourcing efforts frequently fail to elicit logically consistent and grammatical explanations \cite{narang2020wt5}.
Furthermore, a lack of standardized crowdsourcing design has resulted in highly varied datasets, which are hard to compare or combine \cite{tan2021diversity}.

Recent progress in prompting large language models (LLMs) provides a potentially promising alternate to large-scale crowdsourcing. The \emph{in-context learning} paradigm,
wherein powerful language models are prompted in a few-shot manner with just a few examples, has proven surprisingly effective across a range of NLP tasks \cite[\textit{i.a.}]{radford2019language, brown2020language, shin-etal-2020-autoprompt, schick-schutze-2021-exploiting}. 
In this work we ask: can LLMs also generate reliable \emph{explanations}?
In human subjects studies, we find that GPT-3 \cite{brown2020language} can be readily made to generate explanations via prompting, and surprisingly, humans often prefer GPT-3 generated explanations to crowdsourced explanations in existing datasets (\S\ref{sec:analysis}).

Two additional human subjects studies, however, demonstrate that GPT-3-generated explanations still have significant room for improvement along axes such as providing new information (i.e., avoiding repetition) and supporting the label; human subjects found less than half of greedily-decoded GPT-3-generated explanations to be \emph{acceptable} with 100\% agreement. 
To improve upon this, we \emph{re-frame} the role of crowd annotators: instead of asking them to write explanations as in prior work, we (1) repeatedly query GPT-3 to generate multiple candidate explanations for each input instance, and (2) ask crowdworkers to rate the acceptability of each candidate generation. 
After showing that GPT-3 can usually generate an explanation that humans unanimously find acceptable within as few as five queries (\S\ref{ssec:room_for_improvement}), we use a small number of these binary crowdworker judgments to supervise an acceptability filtering model, which can be applied to select high quality candidates among GPT-3's outputs (\autoref{fig:filter_model}; \S\ref{sec:improved_modeling}).

Despite intrinsic subjectivity in acceptability ratings, our supervised model improves upon the already-competitive few-shot paradigm by consistently selecting (human-identified) high quality explanations better than strong baselines.
Human evaluations reveal that the filtration model not only improves acceptability, but also other axes like supporting the label and providing novel information.

In summary, our main findings are:
\begin{enumerate}[leftmargin=*,topsep=0pt,itemsep=-1ex,partopsep=1ex,parsep=1ex]
    \item[i.] few-shot prompting with GPT-3 produces surprisingly competitive explanations, providing a promising alternative to crowd-authored free-text explanation corpora;
    \item[ii.] binary human labeling can instead be leveraged to train a filter that selects high-quality machine-generated explanations; and
    \item[iii.] in areas where GPT-3 struggles, including information content, supporting the label, and overall acceptability, our proposed overgenerate-and-filter pipeline improves generated explanations.
\end{enumerate}
We publicly release our code and data.\footnote{\url{https://github.com/allenai/few_shot_explanations/}}
\section{GPT-3 is Competitive with Crowdsourced Explanation Datasets}
\label{sec:analysis}

We investigate three research questions:
\begin{enumerate}[leftmargin=*,topsep=0pt,itemsep=-1ex,partopsep=1ex,parsep=1ex]
\item Are GPT-3-generated explanations preferable to crowdsourced ones in existing datasets? (\S\ref{ssec:gpt3-vs-human-explanations})
\item Can improving prompt quality improve GPT-3-generated explanations? (\S\ref{ssec:crowdworkers-vs-expert-prompts})
\item Along what fine-grained dimensions are GPT-3-generated explanations preferred, and do these correlate with overall acceptability? (\S\ref{ssec:absolute_analysis})
\end{enumerate}

\paragraph{Explanation tasks and datasets.}
We consider two English tasks: CommonsenseQA and natural language inference (NLI), shown in \autoref{tab:examples}.
CommonsenseQA \cite{talmor-etal-2019-commonsenseqa} is a multiple choice task posed over commonsense questions.
Crowdsourced free-text explanations for instances in CommonsenseQA are provided in the CoS-E v1.11 \cite{rajani-etal-2019-explain} and ECQA \cite{aggarwal-etal-2021-explanations} datasets.
ECQA explanations are counterfactual, i.e., annotators were instructed to explain not only the correct answer choice but also why the others are incorrect.
ECQA was released to address the quality issues of CoS-E \cite{narang2020wt5}; for completeness, we experiment with both.
Our second task is NLI, which involves inferring whether a given hypothesis sentence entails, contradicts, or is neutral towards a premise.
This task is instantiated with the SNLI dataset \cite{bowman-etal-2015-large} and crowdsourced explanations from e-SNLI \cite{camburu2018snli}.
For each task, we report results on a fixed, randomly-sampled 250-instance test set not observed during prompt design. 

\begin{table}[t!]
\centering
\small
\begin{tabular}{p{0.90\linewidth}}
\toprule
\textbf{SNLI \cite{bowman-etal-2015-large}} \\
\midrule[0.03em]
\textbf{Premise:} Dark-haired man wearing a watch and oven mitt about to cook some meat in the kitchen.\\
\textbf{Hypothesis:} A man is cooking something to eat.\\
\textbf{Label:} entailment\\ \\

\textbf{e-SNLI \cite{camburu2018snli}:} Meat is cooked in a kitchen, and is a food that you eat. Using an oven mitt implies you're about to cook with hot utensils.\\
\textbf{GPT-3:} Cooking is usually done to prepare food to eat.
\\
\toprule
\textbf{CommonsenseQA \cite{talmor-etal-2019-commonsenseqa}} \\
\midrule[0.03em]
\textbf{Question:} What is the result of applying for  job? \\
\textbf{Answer Choices:} anxiety and fear, increased workload, praise, less sleep, or being employed\\ 
\textbf{Correct Choice:} being employed \\ \\
\textbf{CoS-E \cite{rajani-etal-2019-explain}:} being employed applying for job\\
\textbf{ECQA \cite{aggarwal-etal-2021-explanations}:} Applying for a job is followed by attending interview which results in being employed. Applying for a job may not result in the other options.\\
\textbf{GPT-3:} Applying for a job can result in being employed, which is a positive outcome.\\
\bottomrule
\end{tabular}
\caption{Task-specific instances, along with their crowdsourced explanations from the respective datasets, shown alongside explanations generated greedily by GPT-3. In our experiments, the SNLI GPT-3 explanation was preferred over its corresponding e-SNLI explanation by $2/3$ annotators. 
For CommonsenseQA, $3/3$ preferred the GPT-3 explanation to the CoS-E explanation, and $2/3$ to the ECQA one. 
}
\label{tab:examples}
\end{table}
\definecolor{question_color}{HTML}{1B9E77}
\definecolor{label_color}{HTML}{D95F02}
\definecolor{explanation_color}{HTML}{7570B3}

\begin{table}[t!]
\centering
\small
\begin{tabular}{p{0.90\linewidth}}
\toprule
Let's explain classification decisions.\\ 
\textcolor{question_color}{A young boy wearing a tank-top is climbing a tree.}\\ 
question: \textcolor{question_color}{A boy was showing off for a girl.}\\ 
true, false, or neither? \textcolor{label_color}{neither}\\ 
why? \textcolor{explanation_color}{A boy might climb a tree to show off for a girl, but he also might do it for fun or for other reasons.}\\ 
\#\#\#\\ 
\textcolor{question_color}{A person on a horse jumps over a broken down airplane.}\\ 
question: \textcolor{question_color}{A person is outdoors, on a horse.}\\ 
true, false, or neither? \textcolor{label_color}{true}\\ 
why? \textcolor{explanation_color}{Horse riding is an activity almost always done outdoors. Additionally, a plane is a large object and is most likely to be found outdoors.}\\ 
\#\#\#\\ 
\textcolor{question_color}{There is a red truck behind the horses.}\\ 
question: \textcolor{question_color}{The horses are becoming suspicious of my apples.}\\ 
true, false, or neither? \textcolor{label_color}{false}\\ 
why? \textcolor{explanation_color}{The presence of a red truck does not imply there are apples, nor does it imply the horses are suspicious.}\\ 
\#\#\#\\ 
\textcolor{question_color}{A dog carries an object in the snow.}\\ 
question: \textcolor{question_color}{A dog is asleep in its dog house.}\\ 
true, false, or neither? \textcolor{label_color}{false}\\ 
why?\\
\bottomrule
\end{tabular}
\caption{Example of a prompt with 3 training examples for SNLI: presented are the \textcolor{question_color}{premise/hypothesis} pairs, the \textcolor{label_color}{gold labels}, and the \textcolor{explanation_color}{explanations} (written by us) that act as input to GPT-3 (in practice, we use 8-24 examples per prompt). The text generated by the model acts as the free-text explanation.
In this case, the model greedily auto-completes (given 12 examples): 
\emph{``A dog cannot carry something while asleep''}.
}
\label{fig:prompt_design}
\end{table}

\paragraph{Few-shot prompting for explanations.}
\label{ssec:in-context}

We use GPT-3 Davinci\footnote{Non-instruct version, available in Summer 2021.} \cite{brown2020language}, an autoregressive language model with \textasciitilde175B parameters trained on a large dataset of text scraped from the internet.
We prompt the model with several (question, gold answer, explanation) triplets followed by an unexplained question-gold answer instance for which we expect the model to generate an explanation.\footnote{We condition on the gold label as a \emph{methodological control} to ensure reliable human evaluation. In pilot studies, we found it hard to avoid bias against explanations when we disagreed with the predicted label. 
Prior work \cite{kayser2021vil, marasovic2021few} has removed this confounder by only considering explanations for \emph{correctly-predicted} instances, which may overestimate explanation quality. Our method allows us to report results on a truly representative sample.
We experiment with incorrect vs. correct predictions in \autoref{appdx:absolute}, finding that GPT-3 can competitively explain gold labels for instances it predicted incorrectly.
}
We use a total of 115 randomly sampled train instances to create our prompts; each prompt consists of 8-24 randomly selected examples from this set.
For each instance, we generate a single explanation with greedy decoding. 
More details about prompt construction are in \autoref{sec:app:priming}; example prompts are given in Tables \ref{fig:prompt_design} and \ref{fig:prompt_design_cose}.

\paragraph{Crowdsourcing evaluation.} 
Given that existing automatic metrics often do not correlate well with human judgements of explanation quality \cite{clinciu-etal-2021-study, kayser2021vil}, we conduct human studies for evaluation.\footnote{
Our studies were conducted on the Amazon Mechanical Turk platform, at the Allen Institute for AI.
We selected workers located in Australia, Canada, New Zealand, the UK, or the US, with a past HIT approval rate of $>$98\% and $>$5000 HITs approved, and compensated them at a rate of \$15/hour.
Each worker completed qualifying exams on explanation evaluation and the NLI task.
} 
We ensure each experiment has a substantial number of distinct crowdworkers to mitigate individual annotator bias (\autoref{table:annotators}).

We present workers with a dataset instance, its gold label, and two explanations for the instance generated under different conditions (``head-to-head''). 
We then ask them to make a preferential selection, collecting 3 annotations per data point.
We report inter-annotator agreement (IAA) using Krippendorff's $\alpha$ \cite{krippendorff2011computing}.
We find low-to-moderate agreement across studies, indicating the subjective nature of the task; see Tables \ref{table:gpt3-vs-crowdsourced-explanations}, \ref{table:prompt_source}, and \ref{table:gpt3-vs-crowdsourced-ourPrompts}.
\autoref{sec:app:crowdsourcing} contains further details on quality control.

\subsection{Are GPT-3 explanations preferred over crowdsourced ones?}
\label{ssec:gpt3-vs-human-explanations}

We perform a head-to-head comparison of explanations generated by GPT-3 with greedy decoding vs. gold human-written explanations in the original datasets. 
The crowdsourced explanations serve as a reasonable upper bound for what a supervised explanation generation model trained on them could produce. 
\autoref{tab:examples} contains examples of GPT-3-preferred explanations.

Results are shown in \autoref{table:gpt3-vs-crowdsourced-explanations}.
GPT-3 greedily-decoded explanations are frequently preferred or comparable to crowdsourced explanations in CoS-E, which is not too surprising given the dataset has many ungrammatical explanations \cite{narang2020wt5}. And, while ECQA and e-SNLI explanations are strongly preferred to GPT-3, there are still a non-trivial number of cases where GPT-3 explanations are competitive (47.3\% and 36.4\%, respectively).

\begin{table}[t]
\centering
\small
\begin{tabular}{lrrrr}
    \toprule
     & \multicolumn{3}{c}{\textbf{Preferred Explanation (\%)}} & \\
     \cmidrule(lr){2-4}
    \textbf{Dataset} & \textbf{Crowd} & \textbf{Tie} &  \textbf{GPT-3} & $\mathbf{\alpha}$\\
    \midrule
    CoS-E & 20.3 & 34.9 & \textbf{44.8} & 0.5 \\
    ECQA & \textbf{52.7} & 12.9 & 34.4 & 0.2 \\
    e-SNLI & \textbf{63.6} & 11.6 & 24.8 & 0.3 \\
    \bottomrule
\end{tabular}
\caption{
Head-to-head human evaluation for 250 explanations generated by GPT-3 vs.\ written by crowdworkers in the datasets, along with Krippendorff's $\alpha$ for IAA. 
Results are shown as \% preferences. 
We prompted GPT-3 with crowd explanations \emph{from the corresponding datasets}.
}
\label{table:gpt3-vs-crowdsourced-explanations}
\end{table}
\begin{figure*}[ht]
	\centering
	\includegraphics[width=2.1\columnwidth]{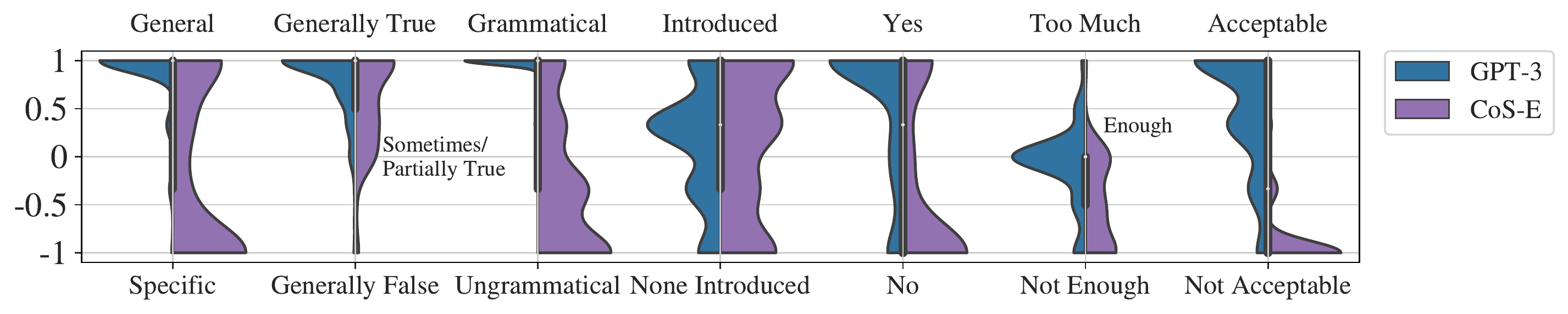}
	\includegraphics[width=2.1\columnwidth]{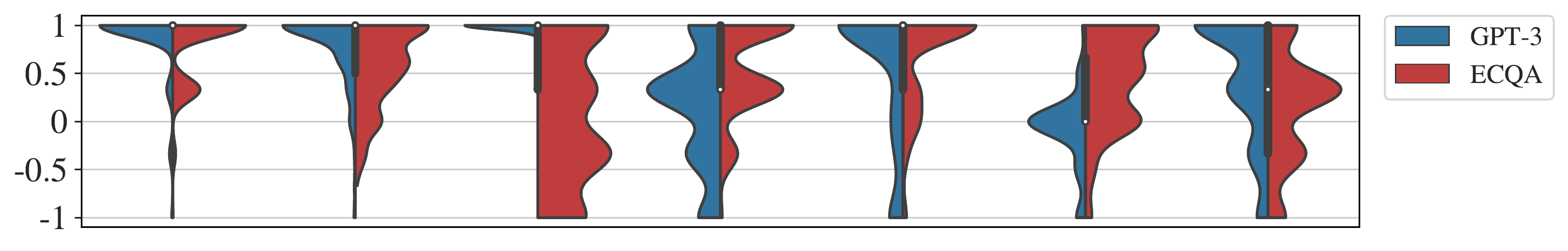}
	\includegraphics[width=2.1\columnwidth]{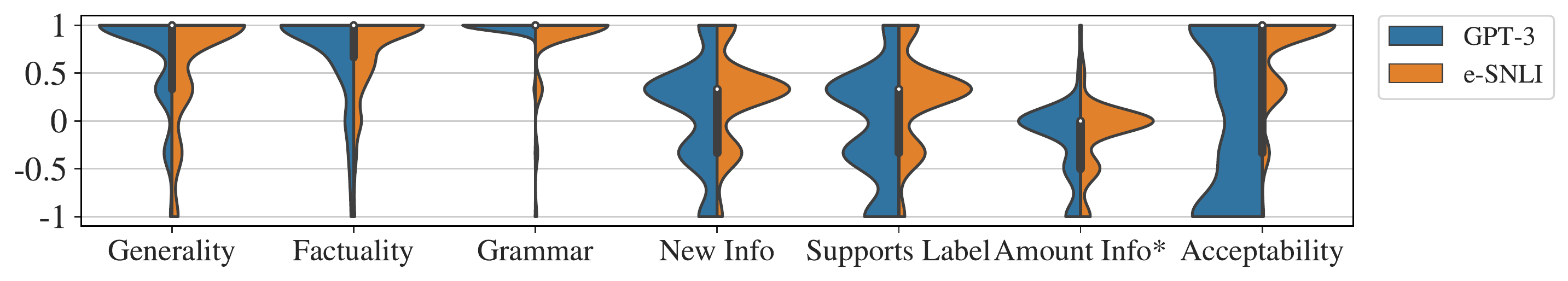}
	\caption{
		Absolute evaluation for GPT-3 and crowdsourced explanations for CommonsenseQA via CoS-E (top) and ECQA (middle) datasets, and for NLI via the e-SNLI dataset (bottom).
		The distribution of mean scores of 3 annotators for each instance in the test set is plotted.
		All attributes besides \textbf{Factuality} and \textbf{Amount Info} are binary. 
		*\textbf{Amount Info} is the only attribute for which a value of 0 is preferred to a value of 1. See \autoref{table:absolute_stats} for more details.
	}
	\label{fig:absolute_all}
\end{figure*}

\subsection{Can improving prompt quality improve GPT-3-generated explanations?}
\label{ssec:crowdworkers-vs-expert-prompts}
\begin{table}[ht]
\centering
\resizebox{\columnwidth}{!}{
\small
\begin{tabular}{lrrrr}
    \toprule
    & \multicolumn{3}{c}{\textbf{Preferred GPT-3 Explanation (\%) }} & \\
    \cmidrule(lr){2-4}
    \textbf{Dataset} & \textbf{Crowd Prompts} & \textbf{Tie} & \textbf{Our Prompts} & $\alpha$\\
    \midrule
     CoS-E & 6.9 & 13.5 & \textbf{79.6} & 0.2\\
     ECQA & 15.9 & 9.5 & \textbf{74.7} & 0.3 \\
     e-SNLI & 30.8 & 26.8 & \textbf{42.4} & 0.5 \\
    \bottomrule
\end{tabular}
}
\caption{Head-to-head human evaluation for 250 explanations generated by GPT-3 prompted with either author-written explanations or crowdsourced explanations from the associated datasets, along with Krippendorff's $\alpha$ for IAA. 
}
\label{table:prompt_source}
\end{table}

Given that low-quality training instances may result in low-quality predictions (especially in a few shot setting),\footnote{For example, GPT-3 reproduces known data artifacts in the CoS-E corpus when prompted with explanations from it, such as the recurring phrase ``rivers flow trough (sp) valleys''.} we ask: can we improve GPT-3 generations simply by conditioning on higher-quality instances?
To this end, we replace the 115 crowdsourced explanations from the original datasets for prompting GPT-3 with explanations carefully written by the authors of this paper (see \autoref{tab:prompt_examples} for examples). 
Our prompts are used to generate a different set of GPT-3 explanations on the same test data. 

We perform a head-to-head human evaluation of the GPT-3 generations conditioned on the explanations we wrote vs. those conditioned on the crowdsourced explanations.
Results in \autoref{table:prompt_source} show that, for all three corpora, generations conditioned on our explanations outperform generations conditioned on crowdsourced ones, illustrating the importance of good-quality prompts for GPT-3. 

We repeat the experiment of \sect{ssec:gpt3-vs-human-explanations}, but with our prompts instead of dataset prompts. With this change, GPT-3 generations are even more competitive (\autoref{table:gpt3-vs-crowdsourced-ourPrompts}).
For all three datasets, more than half the time, few-shot prompting results in an explanation \emph{at least as good as} a human-written explanation.
For subsequent experiments, we prompt GPT-3 with the author-written explanations.

\begin{table}[t]
\centering
\small
\begin{tabular}{lrrrr}
    \toprule
     & \multicolumn{3}{c}{\textbf{Preferred Explanation (\%)}} & \\
     \cmidrule(lr){2-4}
    \textbf{Dataset} & \textbf{Crowd} & \textbf{Tie} &  \textbf{GPT-3} & $\alpha$\\
    \midrule
    CoS-E & 7.2 & 13.9 & \textbf{78.9} & 0.5 \\
    ECQA & 44.5 & 9.7 & \textbf{45.7} & 0.4 \\
    e-SNLI & \textbf{49.6} & 9.7 & 40.7 & 0.2 \\
    \bottomrule
\end{tabular}
\caption{
Head-to-head human evaluation for 250 explanations generated by GPT-3 vs.\ written by crowdworkers in the datasets, along with Krippendorff's $\alpha$ for IAA. 
GPT-3 explanations were prompted with explanations \emph{written by the authors}.
}
\label{table:gpt3-vs-crowdsourced-ourPrompts}
\end{table}

\subsection{What types of explanations does GPT-3 generate?}
\label{ssec:absolute_analysis}

Pairwise evaluations can only offer perspective on the relative quality of generated explanations. Are crowd annotators simply comparing explanations on surface-level features like grammaticality? 

To understand finer-grained characteristics of explanations, we design a second human study to collect absolute Likert-scale judgments across seven axes of quality (with each explanation judged by 3 annotators).
The first three axes capture surface-level features: generality, grammaticality, and factuality. 
The next three capture richer aspects of explanation quality: whether new information is introduced (a requirement for non-vacuous explanations), whether explanations support the gold label, and whether the amount of information given is sufficient. Finally, we ask for an overall judgement of quality: \emph{is the explanation acceptable}?
We explain our design process in Appendix \ref{sec:app:absolute}. Results on the crowdsourced and GPT-3 explanations for both tasks are given in \autoref{fig:absolute_all}.\footnote{Krippendorff's $\alpha$ is 0.48 for CommonsenseQA annotations and 0.31 for SNLI; see \autoref{table:agreement_absolute}.}

For both tasks, GPT-3 explanations do well in all 3 surface-level categories, with statistically significantly greater ratings in generality and grammaticality (and factuality for CommonsenseQA) compared to crowdsourced explanations, and distributional means close to $1$. In these categories, there is little room for improvement.

On the other hand, GPT-3 explanations do not contain as much new information as ECQA and e-SNLI explanations, indicating substantial room for improvement (mean=$0.1$ for both tasks compared to $0.6$ for ECQA and $0.2$ for SNLI; these differences are statistically significant at $p\leq0.01$).
GPT-3 explanations are substantially more supportive of the label vs. CoS-E, but not as supportive as ECQA or e-SNLI (all statistically significant at $p\leq0.1$). Indeed, the mean rating of GPT-3 explanations for label support is $0.5$ for CommonsenseQA and $-0.1$ for NLI, demonstrating room for improvement.
These axes are crucial to ensuring explanations are not vacuous and are on-topic.
Finally, GPT-3 explanations are judged as acceptable at higher rates than CoS-E or ECQA explanations, but not e-SNLI explanations. 
Mean scores of $0.5$ (CommonsenseQA) and $0.0$ (NLI) indicate that GPT-3 explanations have room to improve overall.\footnote{We do not evaluate explanations counterfactually, which may explain why ECQA explanations are often labeled as having ``too much information''. See Appendix \ref{ssec:labels}.} 

\paragraph{Correlation between acceptability and other attributes}
To understand what factors are important for the overall ``acceptability" judgement, we compute Spearman correlation \cite[$\rho$;][]{spearman1987proof} between acceptability and all other attributes (\autoref{tab:correlations}). Each is positively correlated with acceptability, though with varying degrees of magnitude. Acceptability is least correlated with ``new information," and most correlated with grammaticality, generality, and the explanation's support for the label.
Overall, the results indicate that human preference for explanations is not fully explained by any one attribute, and is not limited to surface-level features. 
\begin{table}[t]
\centering
\small
\begin{tabular}{lrr}
    \toprule
    \textbf{Attribute} & \textbf{$\rho$} & \textbf{$n$}\\
    \midrule
    Generality & 0.25 & 3750 \\
    Factuality & 0.17 & 2958 \\
    Grammar & 0.31 & 3750 \\
    New Info & 0.05 & 3750 \\
    Supports Label & 0.22 & 2943 \\
    $^*$Amount Info & 0.16 & 1761 \\
    \bottomrule
\end{tabular}
\caption{
Spearman's correlation ($\rho$) between acceptability and other attributes in \autoref{fig:absolute_all}. 
For \textit{Amount Info}, a value of 0 is preferred to a value of 1. 
All results are statistically significant at $p<$0.01.
}
\label{tab:correlations}
\end{table}
\section{Beyond Greedy Explanations}
\label{ssec:room_for_improvement}

While GPT-3 explanations demonstrate strength across surface-level features and are surprisingly competitive in head-to-head settings, they can still be improved.
One might imagine a setting with multiple end-users in which we wish to provide the most \textbf{unambiguously acceptable} explanation as output from a system.
When considering the data from \S\ref{ssec:absolute_analysis}, we find that only $46.3\%$ of the greedily-decoded GPT-3-generated explanations for CommonsenseQA and $31.5\%$ for NLI are rated acceptable \textbf{by $\mathbf{3/3}$ annotators}.\footnote{In \sect{sec:improved_modeling}, we show that these are not upper-bounds caused by intrinsic subjectivity, and that they can be improved upon.} 

Inspired by work in other generation tasks \cite{holtzman2019curious, massarelli-etal-2020-decoding, holtzman-etal-2021-surface}, we hypothesize that equally or more informative explanations can be generated by sampling stochastically.
We sample 4 additional generations from GPT-3 for each instance to complement the greedy generation.
We crowdsource 3 acceptability annotations for each new explanation.

As expected, sampled explanations exhibit lower $3/3$ acceptability than greedy explanations ($25.1\%$ for CommonsenseQA; $11.3\%$ for SNLI).
However, this results in a surprisingly higher proportion of instances that have \emph{at least one} acceptable explanation in the set of 5. 
The greedy explanation was judged to be $3/3$ acceptable in $46.3\%$ of instances for CommonsenseQA and $31.5\%$ for NLI; this increases to $79.5\%$ and $51.2\%$, respectively, when sampled explanations are included.\footnote{\autoref{sec:app:accept_stats} provides similar results for the $2/3$ acceptability threshold.}

\section{Improving Explanation Generation with Acceptability Filtering}
\label{sec:improved_modeling}

The challenge of overgeneration is that GPT-3 alone cannot discern \emph{which} of its stochastic samples are acceptable. Inspired by \citet{west2021symbolic}, we explore training a supervised filter on the collected labels. 
Our key intuition is that by re-framing the role of annotators from explanation authors to binary judges, we can alleviate the need to collect a large-scale explanations dataset---the result is a simpler, cheaper, and easier crowdsourcing setup to administer (\S\ref{ssec:acceptability_dataset}).
Namely, we can (1) aggregate ratings over multiple annotators to produce more reliable labels, (2) use numerical metrics of annotator agreement to remove annotators contributing noisy data, and (3) collect annotations more quickly and cheaply than asking annotators to hand-write explanations.
Moreover, we find that the filter can be trained with a relatively small amount of binary human judgments (\S\ref{ssec:acceptability-training}). 
We demonstrate that the trained model is not simply taking advantage of surface-level features (\S\ref{ssec:acceptability-results}).
\autoref{fig:filter_model} presents an overview of our pipeline.

\subsection{Acceptability Annotations} 
\label{ssec:acceptability_dataset}

We generate train and validation sets by repeating the procedure of generating 1 greedy and 4 sampled explanations for 
991 and 1K instances, respectively, of the CommonsenseQA and SNLI training sets. Combining these with the annotated test sets from previous experiments results in a dataset of 1241/1250 instances in a 72/8/20\% train/val/test ratio for each task. 
We again collect 3 binary acceptability ratings for each instance, resulting in \textasciitilde$6200$ instance-explanation pairs and \textasciitilde19K individual annotations per task. \autoref{table:agg_accept_stats} contains statistics. 
To ensure that models trained on these corpora do not overfit to specific annotators \cite{geva-etal-2019-modeling}, we collect an additional set of judgments for the test set of SNLI from a group of annotators who did not participate in any of our previous annotation tasks (``Test2''). 
\autoref{fig:acceptability_overview} and \autoref{fig:acceptability_examples} show the user interface.\footnote{Krippendorff's $\alpha$ for all acceptability annotations is 0.34 for CommonsenseQA and 0.39 for SNLI (see \autoref{table:agreement_relative}).}

While we evaluate at test-time with the schema that only instances that $3/3$ annotators deem acceptable are considered acceptable, preliminary experiments show that treating both $2/3$ and $3/3$ agreement instances as acceptable during training performs best on the $3/3$ evaluation criterion at test-time.\footnote{Our results don't significantly change if a $2/3$ cutoff is used at test time instead: \autoref{sec:app:additional_filter_results_2outof3} contains the results.} 
We also train a variant where we randomly select one annotation from the three as the gold label (``without human agreement'').

\subsection{Acceptability Filter}
\label{ssec:acceptability-training}

Concretely, given the problem instance, the gold label, and a generated explanation, the acceptability filter predicts whether the explanation is acceptable.
We fine-tune two sequence-to-sequence architectures, T5-Large \cite{raffel2019exploring} and T0-3B \cite{sanh2021multitask}. 
Each model is trained 5x with different random seeds.
Further training details are given in \autoref{sec:app:training}.

\paragraph{Baselines.}
We train an \textbf{explanation-only} baseline, which receives as input only the explanation; similar baselines have been proposed for NLI \cite{poliak-etal-2018-hypothesis, gururangan-etal-2018-annotation}. 
These models represent the hypothesis that annotator ratings can be reconstructed with only surface features of the explanation candidates, e.g., grammaticality. We also consider a \textbf{negative log-likelihood (NLL)} baseline, which uses GPT-3's estimated probability as the acceptability classification score. This is a slightly more competitive baseline than greedy; greedy usually (but not always) produces the highest-likelihood explanation.\footnote{
According to GPT-3, a sampled explanation from the set of 5 has a lower NLL than greedy for only 2.8\% and 3.6\% of instances, respectively, of our Com.QA and SNLI test sets.}

\subsection{Evaluation}
\label{ssec:acceptability-eval}

We consider three evaluation settings. 
The first is  \textbf{instance-level} (``select-1''), where the system returns 1 explanation selected from the set of 5 for each instance. 
We return the explanation with the highest model-estimated probability and report instance-level accuracy, i.e., the \% of instances for which a gold acceptable explanation is selected.

We also evaluate at the \textbf{explanation-level}, where we treat each explanation independently and compute metrics over the full dataset. This aligns with the binary classification training of the models (cross-entropy on the explanation labels) and is suited for the setting in which we want to return \emph{all} of the acceptable explanations per instance. In this setting, we report \textbf{average precision (AP)}, an estimate of area under the precision-recall curve.

Finally, we perform an absolute human evaluation (\sect{ssec:absolute_analysis}) on the subset of instances where the filter model does \emph{not} select the greedy explanation as the best, i.e., comparing ``select-1'' performance to a greedy baseline on the instances where it differs.
We additionally re-perform the head-to-head comparison of \autoref{table:gpt3-vs-crowdsourced-ourPrompts}, replacing the greedy GPT-3 explanations with those selected by the filter.

\subsection{Results}
\label{ssec:acceptability-results}

\begin{table}[t]
\centering
\small
\resizebox{\columnwidth}{!}{
\begin{tabular}{lllll}
    \toprule
     & \multicolumn{2}{c}{\textbf{``Select-1'' Acc@3/3}} & \multicolumn{2}{c}{\textbf{Expl.-level AP@3/3}}\\
     \cmidrule(lr){2-3}\cmidrule(lr){4-5}
     & \textbf{Dev} & \textbf{Test} & \textbf{Dev} & \textbf{Test}\\
    \midrule
    Random & $26.8_{3.2}$ & $30.6_{2.1}$ & $27.4_{1.1}$ & $31.6_{0.4}$ \\
    Constant & --- & --- & $27.9$ & $31.8$ \\
    NLL  & $41.8$ & $52.0$ & $42.4$ & $45.6$ \\
     \midrule[0.03em]
    T5-L Expl.-only & $40.2_{3.9}$ & $49.8_{1.1}$ & $43.5_{1.5}$ & $50.0_{1.9}$\\
    T0-3B Expl.-only & $42.6_{1.4}$ & $47.3_{1.6}$ & $41.1_{2.0}$ & $54.1_{1.7}$ \\
    
    \midrule[0.03em]
    T5-L w/o HA & $46.6_{2.3}$ & $55.4_{3.2}$ & $47.0_{2.3}$ & $56.9_{3.6}$\\
    T5-L & $46.4_{2.9}$ & $55.4_{2.1}$ & $45.1_{3.4}$ & $58.3_{1.5}$ \\
    T0-3B w/o HA & $48.4_{2.0}$ & $57.4_{2.8}$  & $44.5_{2.3}$ & $59.8_{2.7}$ \\
    T0-3B & $\mathbf{48.6_{0.9}}$ & $\mathbf{59.9_{1.1}}$ & $\mathbf{49.7_{1.6}}$ & $\mathbf{64.0_{1.5}}$ \\
    \midrule
    Oracle U.B. & $78.0$ & $82.0$ & $100.0$ & $100.0$\\
    \bottomrule
\end{tabular}
}
\caption{Results for acceptability classifiers trained on CommonsenseQA. 
Subscripts indicate standard error over models trained with 5 different random seeds. ``w/o HA'' = without human agreement. ``Oracle U.B'' indicates upper bound based on dataset properties (\sect{ssec:room_for_improvement}).
}
\label{table:accept_results_commonsenseqa}
\end{table}
\begin{table}
\centering
\small
\resizebox{\columnwidth}{!}{
\begin{tabular}{lllllll}
    \toprule
    & \multicolumn{3}{c}{\textbf{``Select-1'' Acc@3/3}} & \multicolumn{3}{c}{\textbf{Explanation-level AP@3/3}}\\
    \cmidrule(lr){2-4}\cmidrule(lr){5-7}
    & \textbf{Dev} & \textbf{Test} & \textbf{Test2} & \textbf{Dev} & \textbf{Test} & \textbf{Test2} \\
    \midrule
    Random & $15.2_{0.7}$ & $14.7_{0.1}$ & $13.6_{0.3}$ & $15.0_{0.6}$ & $14.4_{0.3}$ &  $13.8_{0.2}$\\
    Constant & --- & --- & --- & $15.2$ & $14.5$ & $13.7$  \\
    NLL & $33.0$  & $32.0$ & $31.6$ & $29.9$ & $32.7$ & $28.5$\\
    \midrule[0.03em]
    Expl.-only & $30.2_{0.8}$ & $30.9_{2.1}$ & $27.8_{1.8}$ & $30.6_{2.5}$ & $30.6_{1.3}$ & $25.9_{2.3}$\\
    w/o HA & $\mathbf{38.2_{1.9}}$ & $38.5_{1.8}$ & $\mathbf{36.2_{1.4}}$ & $\mathbf{49.3_{5.3}}$ & $\mathbf{48.5_{3.3}}$ &  $\mathbf{52.8_{5.3}}$ \\
    Full & $37.8_{0.8}$ & $\mathbf{38.7_{0.7}}$ & $35.0_{1.2}$ & $46.8_{3.6}$ & $47.6_{3.5}$ &  $49.5_{4.8}$ \\
    \midrule
    Oracle U.B. & $51.0$ & $52.4$ & $46.4$ & $100.0$ & $100.0$ & $100.0$ \\
    \bottomrule
\end{tabular}
}
\caption{Results for SNLI explanation acceptability; all model results are on T0-3B. 
See \autoref{table:accept_results_commonsenseqa}'s caption.}
\label{table:accept_results_snli}
\end{table}

Classifier performance is given in Tables \ref{table:accept_results_commonsenseqa}-\ref{table:accept_results_snli}. 

\paragraph{Effect of model size.}
On CommonsenseQA, T0-3B outperforms T5-Large by \textasciitilde2-4\% select-1 accuracy and \textasciitilde5-6\% explanation-level AP across splits.
We use T0-3B in subsequent experiments.

\begin{figure*}[t]
	\centering
	\includegraphics[width=2.1\columnwidth]{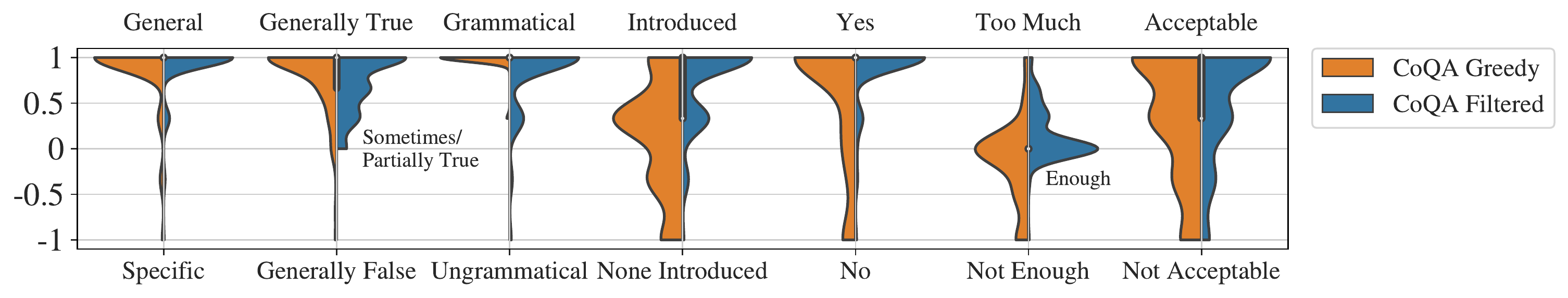}
	\includegraphics[width=2.1\columnwidth]{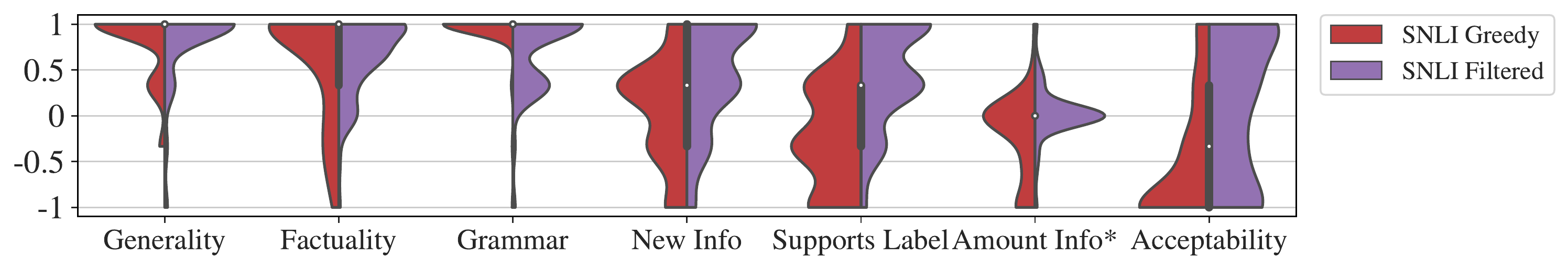}
	\caption{
		Absolute evaluation results in the ``select-1'' setting for the instances where our best-performing filter model does \emph{not} select the greedy explanation (156 instances for CommonsenseQA (top); 91 for NLI (bottom)).
		See caption of \autoref{fig:absolute_all} and the Appendix-\autoref{table:absolute_greedy_vs_filtered_stats} for more details and statistical significance results. 
	}
\label{fig:absolute_greedy_vs_filtered}
\end{figure*}
\begin{table}[t]
\centering
\small
\begin{tabular}{lrrrr}
    \toprule
     & \multicolumn{3}{c}{\textbf{Preferred Explanation (\%)}} & \\
     \cmidrule(lr){2-4}
    \textbf{Dataset} & \textbf{Gold-Standard} & \textbf{Tie} &  \textbf{GPT-3} & $\alpha$\\
    \midrule
    CoS-E & 2.9 & 3.6 & \textbf{93.5} & 0.25 \\
    ECQA & 19.9 & 10.0 & \textbf{70.1} & 0.05 \\
    e-SNLI & \textbf{49.9} & 6.9 & 43.2 & 0.14 \\
    \bottomrule
\end{tabular}
\caption{
A replication of \autoref{table:gpt3-vs-crowdsourced-ourPrompts}, except the GPT-3 explanations are now the top-1 of our filter system. 
}
\label{table:gpt3-vs-filtered}
\end{table}

\paragraph{NLL baseline vs. full model.} 
For both tasks on both validation and test sets, T0-3B outperforms the NLL baseline substantially.
On CommonsenseQA, we observe a 7-8\% gain in instance-level accuracy, and a gain of 18\% explanation-level AP on the test set. 
This provides strong evidence that the supervised model is able to incorporate binary human feedback to predict acceptable explanations at a level much higher that GPT-3 achieves on its own.
We present examples where ``select-1'' predictions differ between NLL and our filter model in \autoref{tab:filter_correct_examples} and \autoref{tab:filter_incorrect_examples}. 

\paragraph{Explanation only vs. full model.} 
Our results suggest that our models are leveraging feature interactions between the instance and explanation to make their predictions: without instance-level context, the explanation-only baselines are on average more than 5 points worse across metrics.
Though they under-perform significantly relative to the full model, explanation-only baselines do fare surprisingly well, indicating that shallow features like factuality and grammaticality may represent latent factors in human acceptability judgments.

\paragraph{The effect of multiple training annotations.} In some cases, performance improves if the training instances are labeled with the consensus of three annotators (vs. the singularly annotated case ``w/o HA"), though the effects are inconsistent. In most cases, using consensus labels results in reduced variance across random seeds. However, the gains may not outweigh the 3x annotations required. 

\paragraph{Our model doesn't overfit to specific annotators.} 
The performance of our model when evaluated on the NLI test set labeled by separate annotators (``Test2'') is comparable to the original test set (instance-level accuracy drops a few points, but explanation-level AP slightly rises).

\begin{table}[t!]
\centering
\small
\begin{tabular}{p{0.90\linewidth}}
\toprule
\textbf{SNLI} \\
\midrule[0.03em]
\textbf{Premise:} An officer in a black uniform and hat stands to the left of a large structure with other officers in the background.\\
\textbf{Hypothesis:} An officer enjoys coffee in a shop.\\
\textbf{Label:} contradiction \\
\textbf{NLL-Predicted Explanation:} An officer in a black uniform and hat is not necessarily an officer enjoying coffee in a shop. \\
\textbf{Filter-Predicted Explanation:} An officer in a structure standing to one side is not the same as enjoying coffee in a shop. \\
\toprule
\textbf{CommonsenseQA} \\
\midrule[0.03em]
\textbf{Question:} Where would there be an auditorium with only a single person speaking? \\
\textbf{Answer choices:} theater, park, \underline{university campus}, crowd, or lights \\
\textbf{NLL-Predicted Explanation:} An auditorium is a large room used for lectures, and a single person speaking is likely to be a lecture.\\
\textbf{Filter-Predicted Explanation:} On university or university-like campuses, the auditoriums are often used for lectures or other University events, where a single person might be speaking.\\
\bottomrule
\end{tabular}
\caption{Randomly-selected instances that our filter model predicts differently than NLL at the ``select-1'' task and got \textbf{correct}, but NLL got incorrect. 
}
\label{tab:filter_correct_examples}
\end{table}

\paragraph{Our model improves generated explanations along desirable traits.} 
We present our absolute human evaluation for greedy vs. filtered explanations from GPT-3 in \autoref{fig:absolute_greedy_vs_filtered}---for both tasks, explanations filtered by our model more readily introduce new information, support the label, and contain at least enough information for both tasks (in addition to being more acceptable).
Interestingly, greedy explanations still prevail in surface-level features (grammaticality and, in the case of CommonsenseQA, factuality; differences are statistically significant with low $p$, see \autoref{table:absolute_greedy_vs_filtered_stats}).\footnote{Krippendorff's $\alpha$ for these experiments is 0.32 for CommonsenseQA and 0.33 for SNLI.}

We additionally find in our head-to-head study (\autoref{table:gpt3-vs-filtered}) that, compared to \autoref{table:gpt3-vs-crowdsourced-ourPrompts}, using filtered GPT-3 explanations instead of greedy increases the preference for GPT-3 explanations by 15-24\% for both CommonsenseQA datasets. We do not see an increase in the SNLI case, which may be due to the fact that fewer GPT-3 explanations change after filtering (36.4\%, compared to 62.4\% for CommonsenseQA), and GPT-3 explanations for SNLI tend to be less acceptable overall, resulting in a lower upper-bound oracle of instances where an acceptable explanation can be selected (\S\ref{ssec:room_for_improvement}). 

\paragraph{In summary.} We have demonstrated the effectiveness of modeling binary crowd judgements of acceptability as a means to select candidates from GPT-3 which are deemed acceptable with unanimous agreement. 
For the method that does not leverage human agreement, this is done with only \textasciitilde5k binary annotations. 
We additionally demonstrate that our filtered explanations improve upon greedy generations in fine-grained categories that probe their topical relevance and meaningful content.
The gap between our best model and the upper-bound oracle indicates that there is still substantial room for improvement in both task settings (but especially for SNLI).
Future work may investigate sampling more explanations, or incorporating other sources of supervision signal.
\section{Related Work}
\label{sec:related}

\paragraph{Free-text explanation generation.}

The earliest neural free-text explanation models did so for computer vision applications \cite{hendricks2016generating, park2018multimodal, kim2018textual} and NLI \cite{camburu2018snli}. 
These methods relied on supervised datasets to train the explanation generator. 
Others have proposed to generate explanations or clarifications to improve task performance in a supervised \cite{rajani-etal-2019-explain, lampinen2022can} or unsupervised \cite{shwartz-etal-2020-unsupervised} manner. \citet{yordanov2021few} study transfer learning between datasets for few-shot generation.

\citet{latcinnik2020explaining} proposed a method to generate free-text explanations supervised only on task signal, and \citet{Brahman2020LearningTR} used sources of weak supervision to generate explanations for defeasible inference. 
\citet{paranjape-etal-2021-prompting} design hand-crafted templates which they use with mask-infilling to produce contrastive explanations from pretrained language models. 

Concurrent work \cite{marasovic2021few} studies the effect of prompt format and model size on crowdworker judgements of prompted explanation plausibility. They find that GPT-3 Davinci outperforms other smaller pretrained models, but that crowdworkers find these explanations less plausible than those from the datasets, aligning with our first experimental result (\autoref{table:gpt3-vs-crowdsourced-explanations}).
We perform a more in-depth study of the fine-grained criteria comprising human acceptability, and demonstrate that with higher-quality prompts and filtering, GPT-3's performance can be significantly improved.

\paragraph{Supervising on human preferences.} 
Prior and concurrent work has used binary judgements from crowdworkers to fit models to human preferences for non-XAI tasks such as summarization \cite{ziegler2019fine, stiennon2020learning}, creating commonsense knowledge bases \cite{west2021symbolic}, and building natural language inference datasets \cite{liu2022wanli}. Unlike these works, we apply human preference modeling to increase the human acceptability of model-generated free-text explanations.
\newcite{west2021symbolic} demonstrate that GPT-3 + a supervised acceptability filter can generate a high-quality causal knowledge graph: in addition to their work being conducted in a different domain, our success conditions and evaluation metrics differ because we \emph{must} produce a prediction for each instance (whereas they can simply discard bad generations).
\section{Conclusion}
\label{sec:discussion}

We demonstrate GPT-3's capacity to generate free-text explanations for NLP task instances in a few-shot setting.
We further improve this capability via an overgenerate + filter approach, where the filter is trained on supervision from human acceptability ratings.
We hope our results can guide future work on free-text explanations via neural or neuro-symbolic systems \cite{Brahman2020LearningTR, majumder2021rationale, saha-etal-2021-explagraphs}. 
Future work may also further investigate the benefits of counterfactual explanations.

While human rationales for decision making are not necessarily the same as model rationales, the goal behind modeling human acceptability is often to build trust with a human user. 
This trust may or may not be warranted \cite{jacovi2021formalizing};
future work would be well-suited to further investigate generated explanations for incorrect label predictions such as in \autoref{appdx:absolute}, which could mislead end users.

\section*{Acknowledgements}

We thank Jena Hwang for helpful advice in designing our human user studies and Peter West for sharing GPT-3 prompting scripts. We thank other members of the Mosaic team at the Allen Institute for AI for valuable suggestions, as well as both our anonymous annotators and reviewers. This work was done while SW was an intern on Mosaic.

\section{Ethics \& Broader Impacts}

All datasets used in this work are public, and we plan to release the machine-generated explanations and annotations we collected. We do not collect any personal information from our human participants.

Models that produce explanations in the means used in our experimental protocol (i.e., by conditioning on the gold labels) have the possibility to cause humans to place unwarranted trust in an AI system. This line of research is complementary to works investigating the faithfulness of model-generated free-text explanations \cite{hase-etal-2020-leakage, wiegreffe-etal-2021-measuring}. We demonstrate in \autoref{ssec:pred_perf} that GPT-3's explanations lack reliability because the model can explain answer choices that were not its prediction equally well. This may be due in part to the fact that decoding algorithms for generating predictions from language models are sub-optimal \cite[e.g.,][]{zhao2021calibrate, holtzman-etal-2021-surface} and GPT-3 may have factual knowledge stored in its parameters about other answer choices that allow it to provide reasonably acceptable explanations. Until this phenomenon is better understood, we do not condone using GPT-3-generated explanations in real-world deployment.

Lastly, our model of human acceptability is based on the aggregate judgements of participants from primarily Western, English-speaking countries working on crowdsourcing platforms. The subjective judgements of explanation acceptability may vary significantly among different population groups.

\bibliography{anthology,custom}
\bibliographystyle{acl_natbib}
\appendix

\section{Prompt Construction}
\label{sec:app:priming}

Following \citet{perez2021true}, we avoid prompt tuning on the full training and development sets of the datasets studied, in order to ensure that our methods represent a true few-shot setting. To develop the initial prompt design, we experimented with no more than 10 different layouts in the GPT-3 Sandbox platform using 15 training examples on the CoS-E and e-SNLI datasets. For subsequent prompt design, we again used these 15 training examples for each dataset from which we sampled 6 prompts, along with a fixed 100-example ``development set'' randomly drawn from the training set. We preserve the ``few-shot'' approach by using a maximum of these same 115 instances to develop our prompting methods. For these 115 examples, the authors of this paper manually wrote high-quality explanations to be used as prompt examples (\autoref{tab:prompt_examples}). As presented in \autoref{fig:prompt_design}, we found that structuring SNLI as a question-answering task achieved the best performance, similarly to \citet{zhao2021calibrate}. We provide an example of our SNLI prompt in \autoref{fig:prompt_design} and CommonsenseQA in \autoref{fig:prompt_design_cose}.

\definecolor{question_color}{HTML}{1B9E77}
\definecolor{label_color}{HTML}{D95F02}
\definecolor{explanation_color}{HTML}{7570B3}

\begin{table}[t!]
\centering
\small
\begin{tabular}{p{0.90\linewidth}}
\toprule
Let's explain classification decisions.\\ 
question: \textcolor{question_color}{When remembering a tragedy in the past, what do many people feel?}\\ 
\textcolor{question_color}{pain, depression, knowing, knowledge, or nostalgia?}\\ 
 \textcolor{label_color}{depression}\\ 
why? \textcolor{explanation_color}{Remembering a past tradedy can resurface feelings that arose in response to that tragedy. Because tragedies are not positive events, it's possible that sadness and depression could arise from remembering it.}\\ 
\#\#\#\\ 
question: \textcolor{question_color}{What do people do sometimes when they find a good deal?}\\ 
\textcolor{question_color}{race cars, murder each other, believe in god, fight each other, or fear death?}\\ 
 \textcolor{label_color}{fight each other}\\ 
why? \textcolor{explanation_color}{Malls sometimes have sales, e.g., on black friday, when they offer good deals; however, the items are sometimes in limited supply, which can cause altercations between folks, each trying to buy the same item.}\\
\#\#\#\\ 
question: \textcolor{question_color}{What does someone who has a greed for energy do?}\\ 
\textcolor{question_color}{buy food, lie, get, cause businesses to grow, or win?}\\ 
 \textcolor{label_color}{buy food}\\ 
why? \textcolor{explanation_color}{When consumed, food provides energy and satisfies the greed for it.}\\ 
\#\#\#\\ 
question: \textcolor{question_color}{Immediately after peeing, a person's bladder is what?}\\ 
\textcolor{question_color}{full, empty, filled, stretchable, or collapsed?}\\ 
 \textcolor{label_color}{empty}\\ 
why?\\
\bottomrule
\end{tabular}
\caption{Example of a prompt with 3 training examples for CommonsenseQA: presented are the \textcolor{question_color}{question} and \textcolor{question_color}{answer choices}, the \textcolor{label_color}{gold labels}, and the \textcolor{explanation_color}{explanations} (written by us) that act as input to GPT-3 (in practice, we use 8-24 examples per prompt). The text generated by the model acts as the free-text explanation.
In this case, the model greedily auto-completes (given 8 examples): 
\emph{``After peeing, the bladder is empty.''}
}
\label{fig:prompt_design_cose}
\end{table}

In-context learning methods have been shown to have high variance based on hyperparameters including example order, number of examples given, and which examples are given \cite{jiang-etal-2020-know, zhao2021calibrate, lu2021fantastically}. While these values have not been standardized, two prominent papers, \citet{schick-schutze-2021-just} and \citet{brown2020language}, use 32 and 64 prompt examples, respectively. Due to the 2049-token limit of the OpenAI GPT-3 API and the fact that the addition of explanations elongates each prompt instance, we find the maximum number of examples the API can accommodate is 24 for CoS-E, e-SNLI, and our handwritten explanations and 16 for ECQA. 

The focus of this work is not on finding the optimal prompt, but on developing a general strategy for few-shot explanation generation that could be successful when no additional (large) validation set for tuning is available. Therefore, to provide as robust of an expected performance estimate as possible, we do not tune the additional hyperparameters, instead sampling them to approximate performance.\footnote{\citet{perez2021true} show that performing cross-validation or tuning via maximum description length over a small validation set does not significantly outperform random sampling of these values.} 
Namely, while prior work uses one fixed prompt for all instances and varies the random seed, 
we approximate the same expected performance by sampling a new set of prompts for each instance. We also sample the number of prompts for each instance (and shuffle their order) from the values $\{8, 16, 24\}$ for CommonsenseQA experiments, $\{8, 16\}$ for experiments using ECQA explanations, and $\{12, 18, 24\}$ for SNLI experiments (to maintain label balance). To overcome label bias in prompt ordering, for tasks with distinct answer choices per instance (CommonsenseQA), we shuffle the answer choices. For tasks with fixed answer choices (SNLI), we sample an equal number of prompt instances for each label (so number of prompt instances is a multiple of 3).

\begin{table}[t!]
\centering
\small
\begin{tabular}{p{0.90\linewidth}}
\toprule
\textbf{SNLI \cite{bowman-etal-2015-large}} \\
\midrule[0.03em]
\textbf{Premise:} A person on a horse jumps over a broken down airplane.\\
\textbf{Hypothesis:} A person is training his horse for a competition.\\
\textbf{Label:} neutral\\
\textbf{Our Explanation:} While it is possible that jumping a horse over an obstacle is part of a training routine for a competition, it is also possible that the horse ride is being done for pleasure, not necessarily for a competion (sp).\\
\textbf{e-SNLI Explanation:} the person is not necessarily training his horse\\
\midrule[0.03em]
\textbf{Premise:} Children smiling and waving at camera\\
\textbf{Hypothesis:} There are children present\\
\textbf{Label:} entailment \\
\textbf{Our Explanation:} Since the children are part of the event of smiling at the camera, they are present at the event under discussion.\\
\textbf{e-SNLI Explanation:} The children must be present to see them smiling and waving.\\
\bottomrule \\
\toprule
\textbf{CommonsenseQA \cite{talmor-etal-2019-commonsenseqa}} \\
\midrule[0.03em]
\textbf{Question:} A cat can't talk, but a cat can what? \\
\textbf{Answer choices:} sleep all day, \underline{meow}, shed fur, see king, live many years \\
\textbf{Our Explanation:} A cat can meow as a way to vocalize.\\
\textbf{CoS-E Explanation:} the cat is a small carnivorous mammal\\
\textbf{ECQA Explanation:} A cat can meow but cannot see the king. Meowing is how a cat communicates and not by sleeping all day, shedding fur or by living many years.\\
\midrule[0.03em]
\textbf{Question:} "There are 10 apples on an apple tree.  Three fall off.  Now there are X apples."  What is this an example of? \\
\textbf{Answer choices:} park, coloring book, garden center, \underline{math problem}, gravity \\
\textbf{Our Explanation:} A math problem is usually posed as a question that requires some operation such as subtraction or addition to answer.\\
\textbf{CoS-E Explanation:} webmath is designed to help you solve\\
\textbf{ECQA Explanation:} Math problem is an arithmetical problem of addition, subtraction, multiplication or division. So ``There are 10 apples on an apple tree.  Three fall off.  Now there are X apples.'' is a math problem. All the other options aren't problems to be examples of the given question.\\
\bottomrule
\end{tabular}
\caption{Examples of explanations used as prompts from various sources, including our handwritten explanations. Correct answers for CommonsenseQA are underlined.}
\label{tab:prompt_examples}
\end{table}

\autoref{tab:prompt_examples} shows a few non-cherry-picked examples of our handwritten explanations used as prompts relative to the datasets.
\section{Crowdsourcing Details}
\label{sec:app:crowdsourcing}

We discuss shared details of the study designs in \S\ref{ssec:labels}. We discuss the head-to-head interface in \S\ref{ssec:head-to-head}, the absolute interface in \S\ref{sec:app:absolute}, and the acceptability interface in \S\ref{sec:app:acceptability}. Finally, we present details on quality control and payment in \S\ref{sec:app:quality_control} and annotator statistics in \S\ref{sec:app:crowd_details}.

\subsection{Shared Interface Details}\label{ssec:labels} For all three human subjects study designs designs, we show the user the input instance (e.g., premise$+$hypothesis) and the gold label in addition to the explanation(s). We explain our motivation for using the gold label as a methodological control in \S\ref{sec:analysis}. 

For a similar reason, we do not show the other incorrect label choices to the user, which is particularly of note for the CommonsenseQA task which has different answer choices per instance. Some instances in CommonsenseQA have multiple correct or very similar answer choices, due to noise in the dataset and the fact that the wrong answer choices were deliberately collected to make the task challenging. We (the authors) again found we struggled to accurately judge explanation quality when we disagreed with the selected answer choice or found multiple answer choices to be correct. To remove this possible confounder, we instruct participants to pretend the gold label is correct even if they disagree with it, and make this easier by hiding the other answer choices. This may result in a slight bias in judgements against the ECQA dataset due to its unique counterfactual nature, though our goal was not to study the benefits and downsides of counterfactual explanations in this work. 

\subsection{Head-to-Head Interface Details}\label{ssec:head-to-head} 
\begin{figure*}
    \centering
    \includegraphics[height=0.8\textheight]{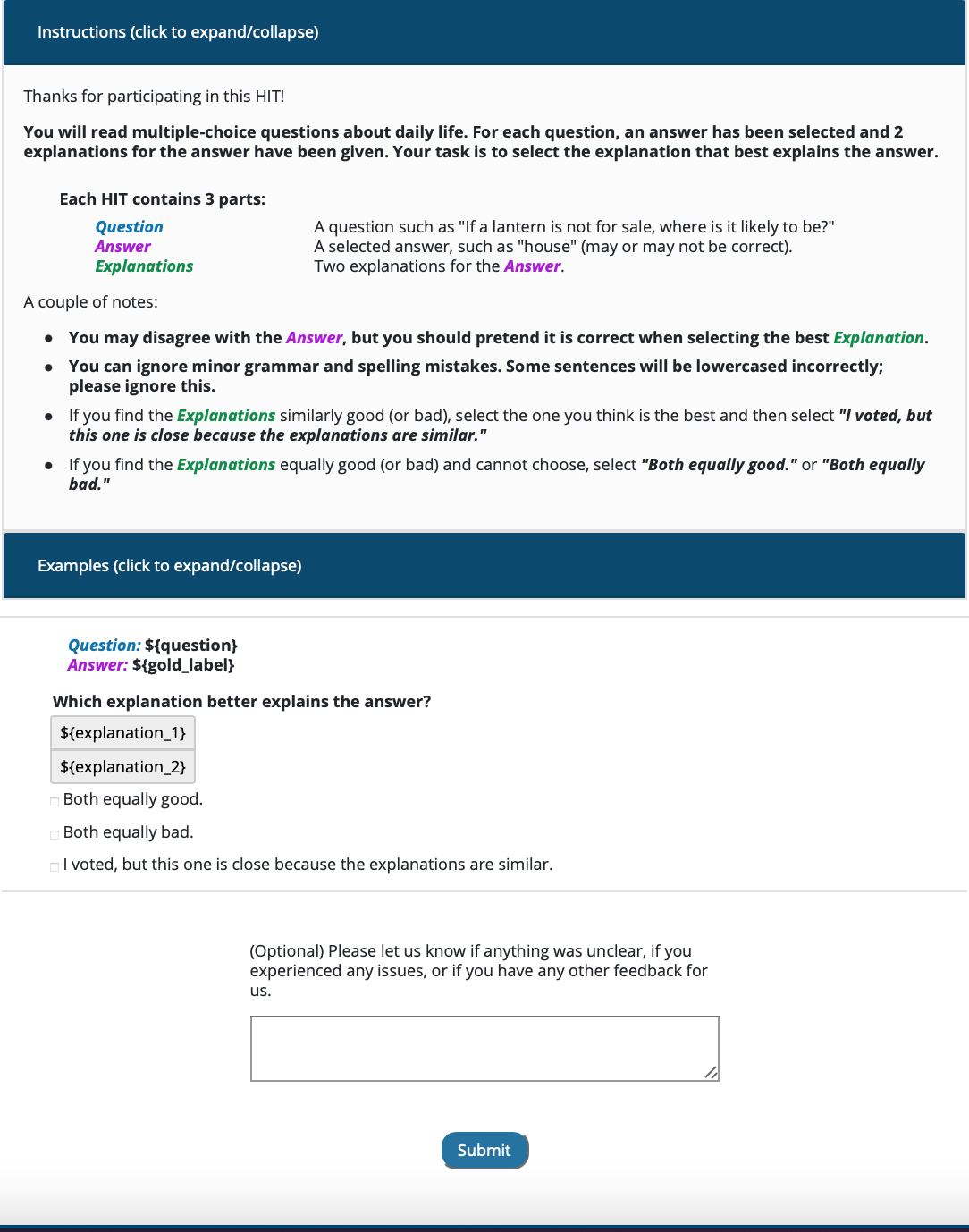}
    \caption{An overview of the user interface of our head-to-head comparison AMT studies for CommonsenseQA. The top shows the instructions and the bottom the actual task. The Examples tab is collapsed here; shown in full in \autoref{fig:relative_examples}.}
    \label{fig:relative_overview}
\end{figure*}

\begin{figure*}
    \centering
    \includegraphics[height=0.75\textheight]{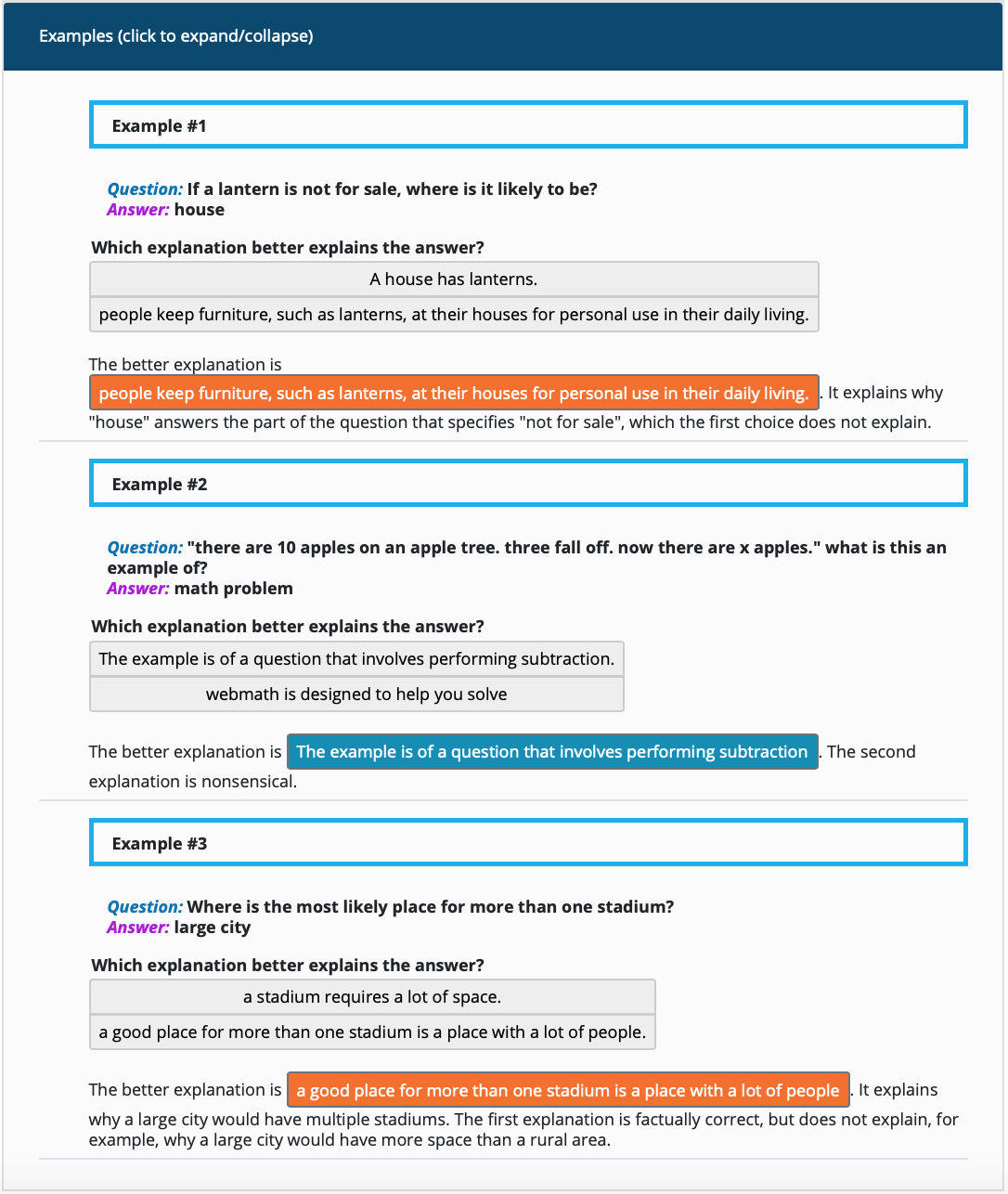}
    \caption{The Examples tab given in the user interface of our head-to-head comparison AMT studies for CommonsenseQA. The full interface is shown in \autoref{fig:relative_overview}.}
    \label{fig:relative_examples}
\end{figure*}

We show the user the task input and gold label, and ask them to select which of two explanations best explains the answer. We instruct workers to consider the gold label to be correct even if they disagree with it (CommonsenseQA instances can be subjective) and to ignore minor grammar and spelling mistakes such as improper upper-casing. Figures \ref{fig:relative_overview} and \ref{fig:relative_examples} show the evaluation interface.

\subsection{Absolute Interface Details}
\label{sec:app:absolute}

\begin{figure*}
    \centering
    \includegraphics[height=0.75\textheight]{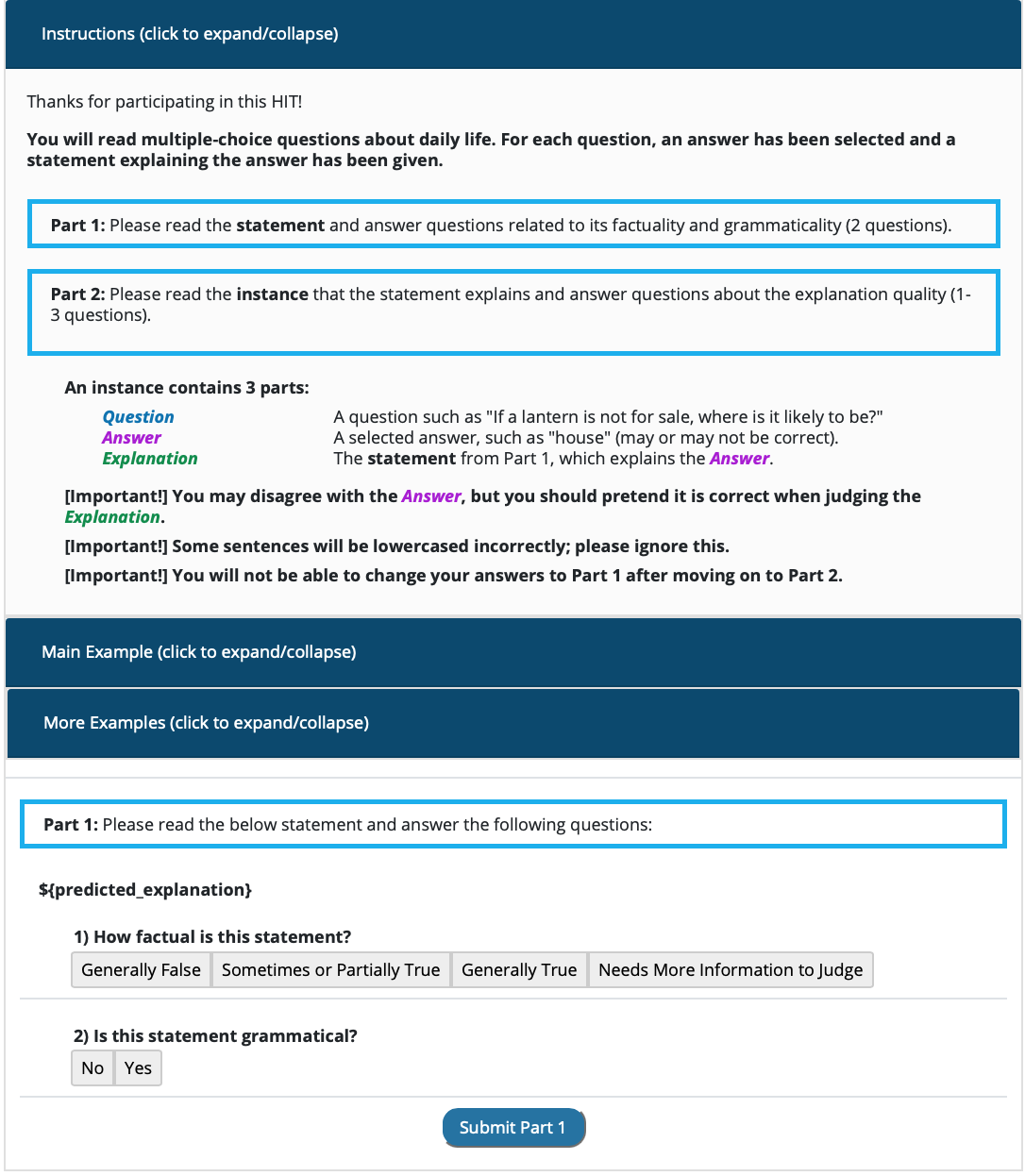}
    \caption{An overview of the user interface template of our absolute comparison AMT studies for CommonsenseQA. The top shows the instructions and the bottom the actual task. Only part 1 of the task is shown here (part 2 appears once part 1 is submitted). The Main Example and More Examples tabs illustrating both parts 1 and 2 are collapsed here; see \autoref{fig:absolute_examples}. }
    \label{fig:absolute_overview}
\end{figure*}

\begin{figure*}
    \centering
    \includegraphics[height=0.8\textheight]{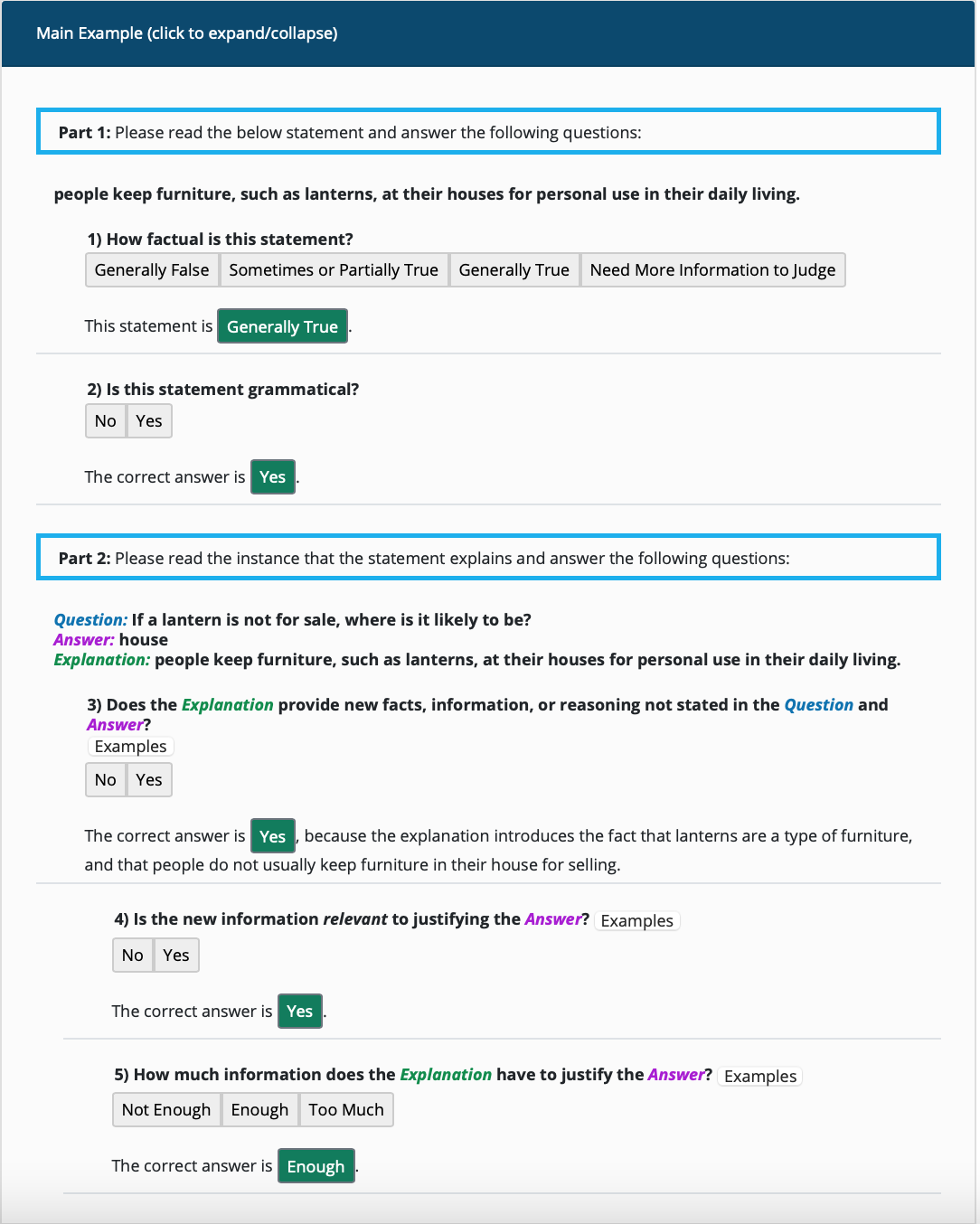}
    \caption{The Main Example given in the user interface template of our absolute comparison AMT studies for CommonsenseQA. This format follows the actual task layout. The full interface is shown in \autoref{fig:absolute_overview}.}
    \label{fig:absolute_examples}
\end{figure*}

Figures \ref{fig:absolute_overview} and \ref{fig:absolute_examples} show the absolute evaluation interface (minus the acceptability attribute, which is collected in a separate run of the study). Our interface is inspired by prior work from psychology and the social sciences \cite{leake1991goal, gopnik1998explanation, lombrozo2007simplicity, zemla2017evaluating, chiyah-garcia-etal-2018-explainable, clinciu-etal-2021-study, sulik2021explanations}.  We iterated over 3-4 versions of the questions and UI design until we had optimized agreement rates as much as possible. Our resulting two-part evaluation consists of 7 questions:

\paragraph{Part 1: Context-Independent Evaluation}
We first assess the explanation in isolation, i.e., these questions are presented to the user without revealing the question/context that the explanation is attempting to address:
\begin{enumerate}[leftmargin=*,topsep=0pt,itemsep=-1ex,partopsep=1ex,parsep=1ex]
    \item \emph{How factual is this statement?} (generally false, sometimes or partially true, generally true, or need more information to judge).  This question is designed to test both generality (can the explanation's truthfulness be ascertained or is more information needed?) and factuality, which aligns with ``compatibility with receiver's existing beliefs'' and that the best explanation is the ``most likely'' explanation from the receiver/user's perspective \cite{lombrozo2007simplicity, zemla2017evaluating, sulik2021explanations}. Generality is coded based on whether a truthfulness answer is selected (considered to be general) or whether the ``need more information to judge'' choice is selected (considered not to be general).
    \item \emph{Is this statement grammatical?} (yes or no) This question is designed to test for clarity, aligning with characteristics such as coherence \cite{lei-etal-2016-rationalizing} and human-likeness and understandability \cite{ehsan2019automated}.
\end{enumerate}

\paragraph{Part 2: Context-Dependent Evaluation}

We next show the user the  question (premise and hypothesis for SNLI) and gold answer that the explanation was conditioned on. We then ask: 

\begin{enumerate}[leftmargin=*,topsep=0pt,itemsep=-1ex,partopsep=1ex,parsep=1ex]
    \item \emph{Does the explanation provide new facts, information or reasoning not stated in the question and answer?} (yes or no) In our preliminary experiments, we found some explanations simply restate the question declaratively with the answer filled in. This question addresses the distinction between ``validity'' and ``utility'' \cite{leake1991goal}: an explanation can be valid (i.e., a restatement of the question with the answer filled-in might be correct), but not useful; utility is defined by whether an explanation ``satisfies an explainer's need for information''. And while utility is best understood in the context of real-world applications \cite{lai2020harnessing}, we nonetheless aim to identify vacuous explanations that do not provide new information.
    \item \emph{Is the new information relevant to justifying the answer? (yes or no)}
    New information, if provided, ``should be compatible with our existing beliefs, and consistent with the evidence and with itself'' \cite{zemla2017evaluating}.
    This question is designed to test whether the information provided supports the label.
    The specific interpretation of ``relevance'' is purposefully left to the annotator.\footnote{This decision is inspired by prior work in psychology, which finds that explanations are only good  ``to the extent that people find [them] satisfying'' \cite{gopnik1998explanation, sulik2021explanations}.}
    \item \emph{How much information does the explanation have to justify the answer?} (not enough, enough, or too much)
    This question is designed to test the extent to which the provided novel information is \emph{adequate} or \emph{sufficient} \cite{kim2016examples, lei-etal-2016-rationalizing, ehsan2019automated}.\footnote{In practice, we do not find Turkers use the ``too much information'' option often, except in the case of ECQA dataset explanations. We included the option because succinctness is an oft-cited explanatory virtue \cite{lombrozo2007simplicity, zemla2017evaluating,  chiyah-garcia-etal-2018-explainable}.}
    \item \emph{Is the explanation acceptable?} (yes or no) The final question is designed to assess annotators' overall judgement of the explanation as a whole.
\end{enumerate}
We only ask Question 2 if the answer to Question 1 is ``yes'' and Question 3 if the answer to Question 2 is yes, because they regard the new facts, information, or reasoning. We found that most prior work tends to lump added-value, relevance, and adequacy judgements into one ``informativeness'' judgement \cite{clinciu-etal-2021-study}, which we felt was too course to allow for meaningful error analysis.

\subsection{Acceptability Interface Details}\label{sec:app:acceptability}

\begin{figure*}
    \centering
    \includegraphics[height=0.8\textheight]{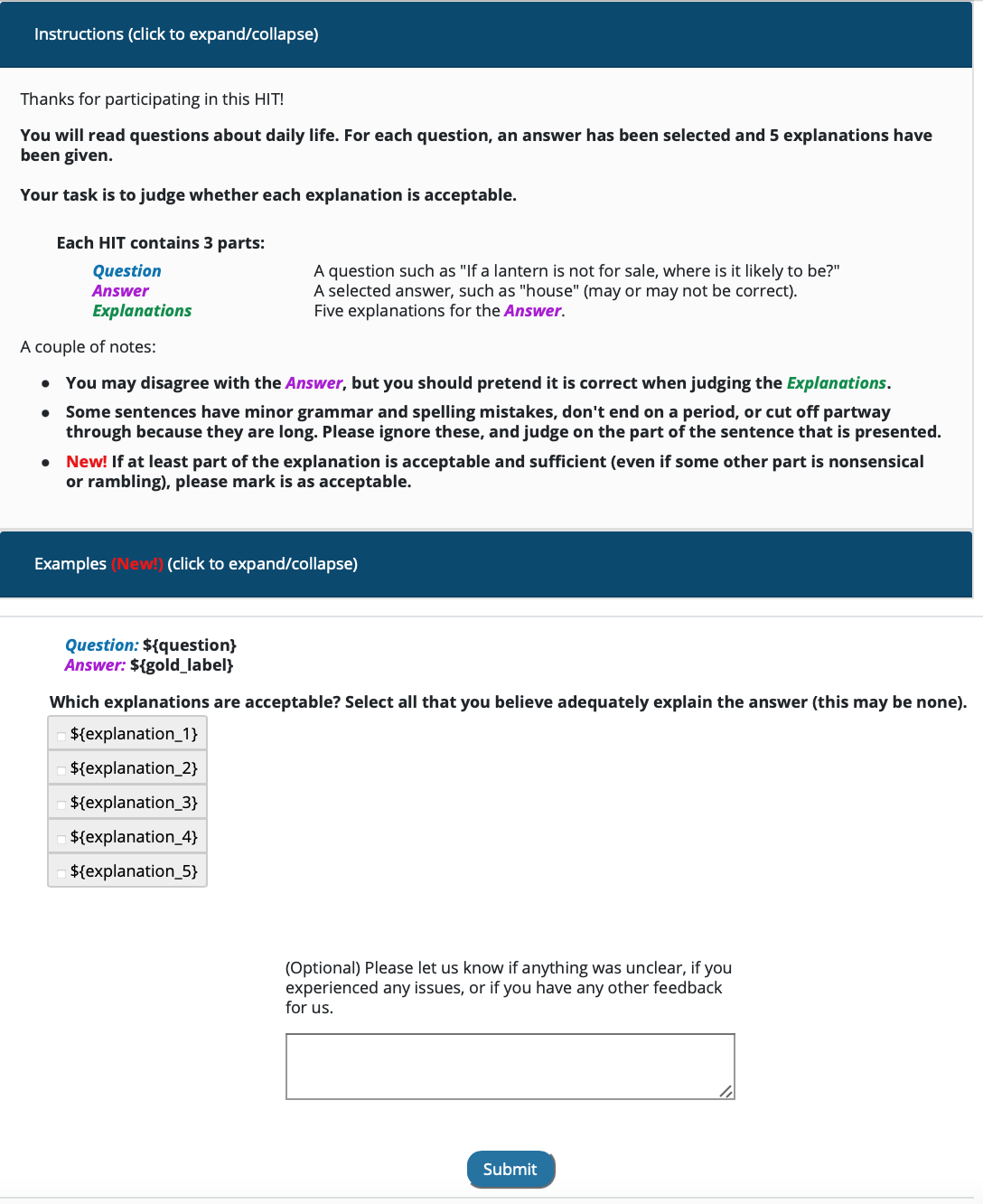}
    \caption{An overview of the user interface of our explanation acceptability AMT studies for CommonsenseQA. 
    The top shows the instructions and the bottom the actual task. 
    The "examples" tab is collapsed here; shown in full in \autoref{fig:acceptability_examples}.}
    \label{fig:acceptability_overview}
\end{figure*}

\begin{figure*}
    \centering
    \includegraphics[height=0.51\textheight]{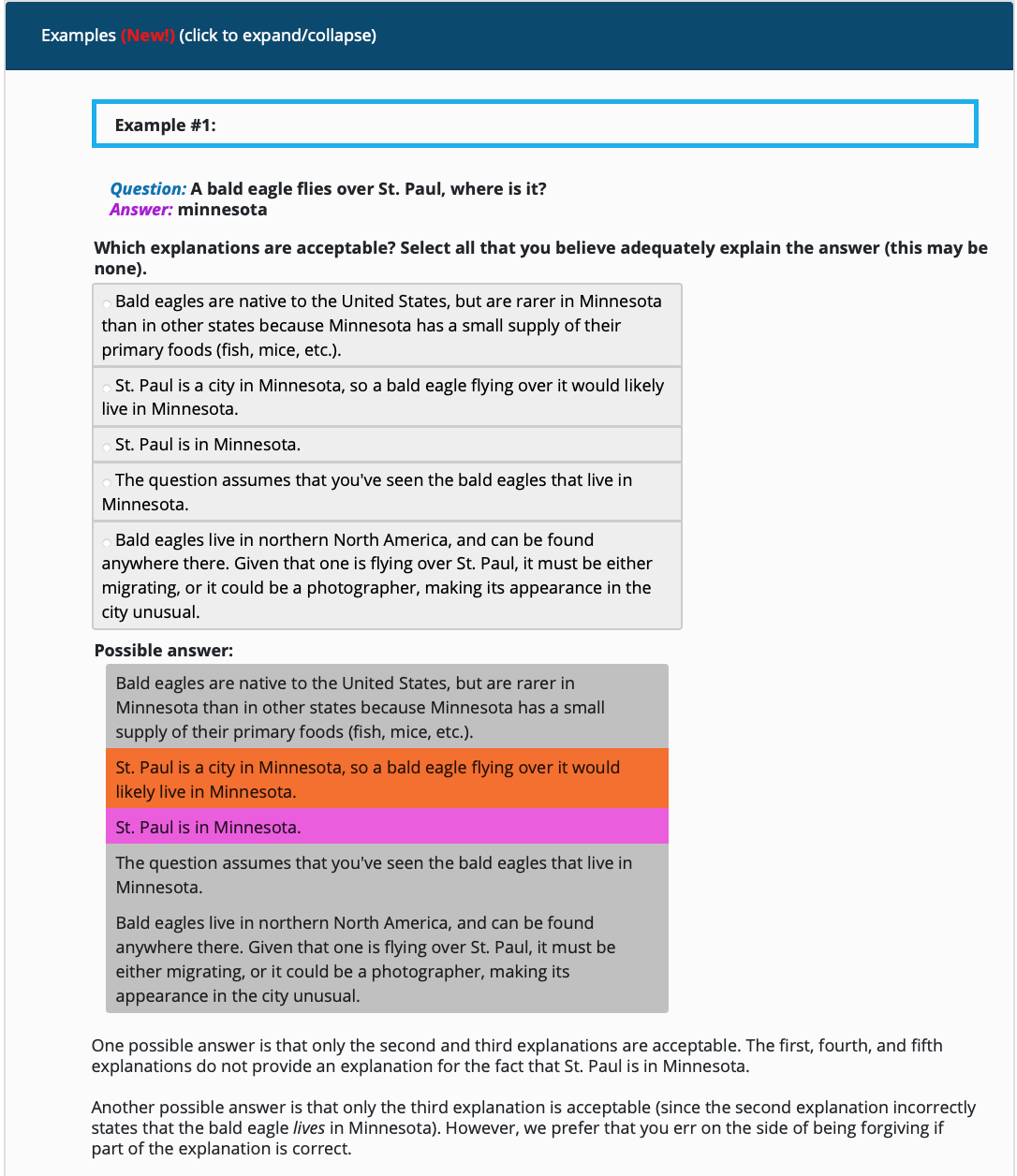}
    \includegraphics[height=0.45\textheight]{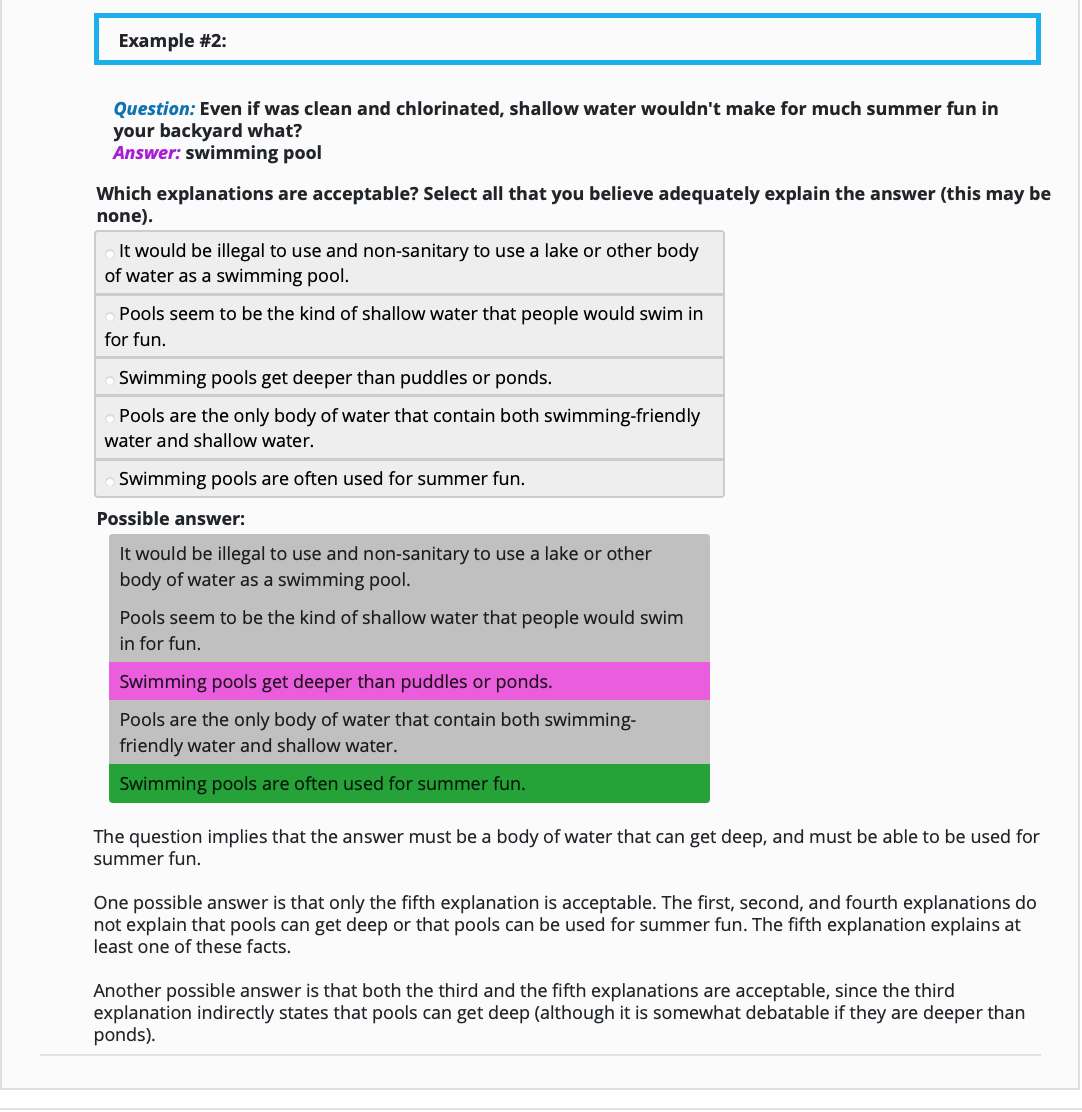}
    \caption{The examples given in the user interface of our explanation acceptability AMT studies for CommonsenseQA. 
    The full interface is shown in \autoref{fig:acceptability_overview}.}
    \label{fig:acceptability_examples}
\end{figure*}

Figures \ref{fig:acceptability_overview} and \ref{fig:acceptability_examples} show the binary acceptability interface used to collect training and test data for the overgeneration filter model.

Spearman's rank-order correlation coefficients (\autoref{tab:correlations}) are computed using scipy \cite{virtanen2020scipy} on the 250 test explanations from the 5 data sources in \autoref{fig:absolute_all}. Each instance is annotated by 3 annotators for a total of 3750 datapoints (some criteria are only evaluated conditionally, resulting in less total annotations-- see Appendix \ref{sec:app:absolute}). Statistical significance is computed using the built-in two-sided non-parametric test. 

\subsection{Quality Control and Payment}
\label{sec:app:quality_control}

We use Amazon Mechanical Turk (AMT), and calculate pay on a rate of \$15/hour. Every few batches, we check to ensure that the median time taken per-annotator amounts to approximately this pay rate. While annotators do tend to speed up the more HITs we released, first-round median times were approximately 30 seconds per head-to-head evaluation HIT (thus paid at \$0.12 each), 1 minute per absolute evaluation HIT (thus paid at \$0.25 each), and 35-39 seconds per acceptability HIT (5 explanations; paid at \$0.20 each).

We require annotators to be located in either Australia, Canada, New Zealand, the United Kingdom, or the United States, as a proxy for English competency.\footnote{We realize this is a broad assumption and likely sub-optimal. However, colleagues have found that broadening the geographical requirements often still leads to $>$90\% of annotators in the US or Canada, due to AMT's pay structure being optimal in these countries.} We require a past HIT approval rate of $>$ 98\% and $>$ 5000 HITs approved. We do not allow annotators to participate who were previously on a block list from our past AMT studies. 

Annotators must complete a qualifying exam in order to participate in the main annotation tasks. The qualifying exam consists of 3 HITs in the same format as the main absolute evaluation task for CommonsenseQA.
We pay \$2.25 for the qualifying exam.
There are 9-18 questions in total (3-6 questions per HIT), some of which are only answerable conditioned on previous answers. A user who answers ``no'' to question 3, for example, will not be asked to answer questions 4 and 5. Given the challenging and sometimes ambiguous nature of some of the questions, for the first run of the qualification exam, we manually awarded qualifications by inspecting the annotators' answers.
Scores for the first run compared to our answers (out of 17 annotators attempting) ranged from 5 to 14 out of 18. The median accuracy was 11 out of 18, and we find that awarding the qualification to those with scores at or above the median aligns closely with our manual inspection. We thus use this score to assign qualifications in future iterations.

Because it is necessary that annotators understand the task before they can evaluate explanation quality \cite{wiegreffe2021teach}, for tasks that are more difficult, i.e., NLI, we additionally require annotators to pass (score of 7 or above) a task-specific qualification exam with 8 questions, paid at \$1.25.

In order to track quality throughout evaluation, we compute inter-annotator agreement using Krippendorff's $\alpha$ and use a hidden built-in Javascript function to compute time per HIT spent. If any annotator completed the tasks in an unreasonably low time, or removing their annotations substantially improves Krippendorff's $\alpha$, we remove their annotations and re-annotate their instances. We additionally ensure that each experiment has a substantial number of distinct crowdworkers to mitigate individual annotator bias, reporting this as well as the mean and median number of HITs completed by each in \autoref{table:annotators}.

\subsection{Statistics}
\label{sec:app:crowd_details}

The number of distinct crowd annotators and the median and mean number of HITs completed 
for each experiment can be found in \autoref{table:annotators}. More detailed breakdowns of inter-annotator agreement for some experiments are in Tables \ref{table:agreement_relative} and \ref{table:agreement_absolute}.
\section{Absolute Evaluation by Label Accuracy}
\label{appdx:absolute}
\label{ssec:pred_perf}

Can GPT-3 produce convincing explanations even for instances it cannot predict correctly?
This has implications for model-generated explanations being ``right for the right reasons''. To produce label predictions, we follow the same prompt format as in Tables \ref{fig:prompt_design} and \ref{fig:prompt_design_cose}, removing the \textsc{why?} token and the gold explanations so that the model generates a label prediction instead.
GPT-3 achieves 50.8\% accuracy on CommonsenseQA compared to a 20\% random baseline, and 46\% accuracy on SNLI compared to a 33.33\% random baseline.\footnote{Low SNLI performance aligns with previous findings that GPT-3 struggles with sentence-comparison tasks \cite{brown2020language, zhao2021calibrate}.}

\begin{figure*}[t]
\centering
   \includegraphics[width=2.1\columnwidth]{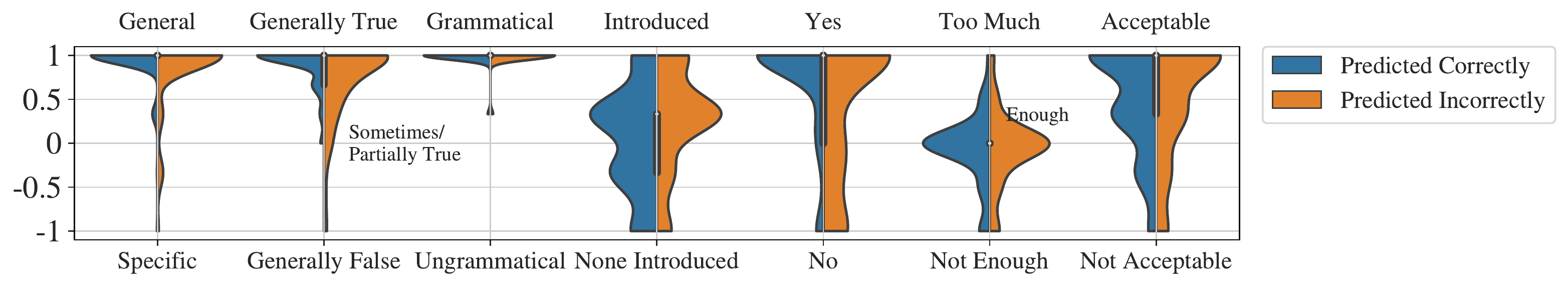}
   \includegraphics[width=2.1\columnwidth]{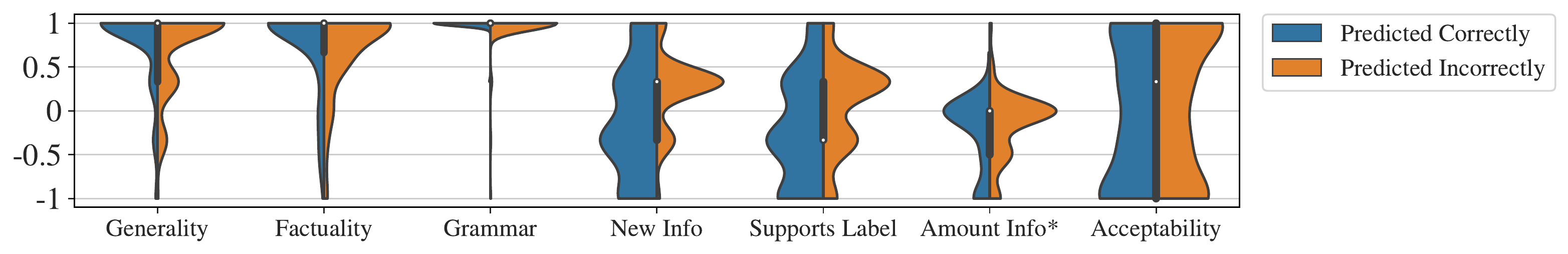}
\caption{Absolute evaluation results for explanations generated by GPT-3 based on whether GPT-3 predicted the instance label correctly or not. CommonsenseQA (top) and SNLI (bottom). See caption of \autoref{fig:absolute_all} and the Appendix-\autoref{table:pc_pu_stats} for more details.
}
\label{fig:absolute_pc_vs_pu}
\end{figure*}

\autoref{fig:absolute_pc_vs_pu} presents absolute evaluation results broken down by whether GPT-3 correctly predicts the instance label. The results show little variation between the correctly-predicted and incorrectly-predicted groups in most attributes.
This indicates that GPT-3 explanations are not faithful enough to use in real-world applications in their current form. 

\section{$\mathbf{2/3}$ Acceptability Statistics}
\label{sec:app:accept_stats}

When we treat explanations rated by at least $2/3$ annotators as ``acceptable'', for CommonsenseQA, $77.9\%$ of greedily-decoded explanations are acceptable; for SNLI, $51.0\%$. $50.5\%$ of sampled explanations are acceptable; for SNLI, $23.5\%$. Out of the set of 5 (1 greedy + 4 stochastic), $97.7\%$ of CommonsenseQA instances have \emph{at least one} acceptable explanation, and $79.5\%$ of SNLI.

\section{Filter Model Details}
\label{sec:app:training}

\begin{table}
\centering
\resizebox{\columnwidth}{!}{
\begin{tabular}{llrrrrr}
    \toprule
    \textbf{Dataset} & \textbf{Split} & \multicolumn{4}{c}{\textbf{\# Instances by Agreement}} & \textbf{Total}\\
    & & \textbf{0/3} & \textbf{1/3} & \textbf{2/3} & \textbf{3/3} & \\
    \midrule
    Com.QA & Train & 932 & 1078 & 1194 & 1296 & 4500\\
    & Dev & 105 & 91 & 132 & 127 & 455\\
     & Test & 298 & 227 & 328 & 397 & 1250\\
    \midrule
    SNLI & Train & 2372 & 805 & 621 & 702 & 4500\\
    & Dev & 272 & 87 & 65 & 76 & 500\\
     & Test & 678 & 225 & 166 & 181 & 1250\\
     & Test2 & 666 & 234 & 179 & 171 & 1250\\
    \bottomrule
\end{tabular}
}
\caption{
Statistics of our acceptability annotations. 
}
\label{table:agg_accept_stats}
\end{table}

We split the 4,955 distinct annotated explanations for CommonsenseQA (5000 for SNLI) into a train/dev set of 4500/455 (4500/500 for SNLI), where all 5 explanations for a given instance are placed in the same set to avoid leakage. We present statistics on the label distribution in \autoref{table:agg_accept_stats}. Along with the metric settings reported in the paper (``select-1'' and explanation-level), we computed a metric that is instance-level but considers all explanations by computing metrics over the 5 explanations of an instance and then averaging across instances, finding in practice that the results are highly similar to the explanation-level evaluation.

We use Huggingface Datasets \cite{lhoest-etal-2021-datasets} and Huggingface Transformers \cite{wolf-etal-2020-transformers} for implementation. We format inputs to the models as follows, where \texttt{expl} is one of the five explanations and the \texttt{gold\_label} is either 0 (not acceptable) or 1 (acceptable):

{\footnotesize
\begin{verbatim}
if explanation_only:
  input_string = (f"explanation: {expl}.
    Is this explanation good or bad?")
else:
  input_string = (
    "{question} answer: {gold_label}. "
    "explanation: {expl}. "
    Is this explanation good or bad?") 
\end{verbatim}
}

The T5-Large model is trained using a learning rate of $1E-4$ with linear decay, a batch size of $64$, and default values for Adam \cite{kingma2014adam}, gradient clipping, and dropout. We train for a maximum $200$ epochs, performing early stopping on the validation loss with a patience of $10$ epochs. For T0-3B, we train with a batch size of $50$. We use AdaFactor \cite{shazeer2018adafactor} with a linear warmup of 500 steps. We conduct an initial hyperparameter sweep over learning rate, considering $1E-5,5E-05,5E-06$. The learning rate that achieves the best validation loss for the full-information model and the explanation-only model is $1E-5$, which we use for all training experiments.

\begin{table}
\centering
\resizebox{0.48\textwidth}{!}{
\begin{tabular}{llc}
    \toprule
    \textbf{Dataset} & \textbf{Split} & \textbf{Krippendorff's $\alpha$}\\
    \midrule
    CommonsenseQA & Training + Validation & 0.32\\
    & Test & 0.40 \\
    \hline
    SNLI & Training + Validation &  0.51 \\
    & Test & 0.50 \\
    & Test2 & 0.47 \\
    \bottomrule
\end{tabular}
}
\caption{Inter-annotator agreement for acceptability AMT studies. 
}
\label{table:agreement_relative}
\end{table}

\begin{table*}
\centering
\resizebox{\textwidth}{!}{
\begin{tabular}{llcccccccc}
    \toprule
    \textbf{AMT Study} & \textbf{Dataset} & \textbf{Generality} & \textbf{Factuality} & \textbf{Grammar} & \textbf{New Info} & \textbf{Supports Label} & \textbf{Amount Info} & \textbf{Acceptability} & \textbf{Aggregate}\\
    \midrule
    GPT-3 Greedy & Com.QA & 0.37 & 0.32 & -0.01 & 0.09 & 0.45 & 0.21 & 0.28 & \textbf{0.38}\\
    GPT-3 Greedy & SNLI & 0.25 & 0.57 & 0.39 & -0.01 & 0.04 & 0.17 & 0.52 & \textbf{0.40}\\
    \hline
    Dataset & CoS-E & 0.71 & 0.38 & 0.36 & 0.42 & 0.68 & 0.08 & 0.25 & \textbf{0.59}\\
    Dataset & ECQA  & 0.01 & 0.21 & 0.30 & 0.00 & 0.03 & 0.25 & 0.04 & \textbf{0.20}\\
    Dataset & e-SNLI & 0.37 & 0.23 & 0.37 & -0.14 & -0.12 & 0.04 & 0.15 & \textbf{0.19}\\
    \hline
    GPT-3 Filtered & Com.QA & 0.25 & 0.18 & 0.25 & 0.11 & 0.27 & 0.15 & 0.30 & \textbf{0.32}\\
    GPT-3 Filtered & SNLI & 0.41 & 0.19 & 0.07 & 0.17 & 0.08 & 0.13 & 0.50 & \textbf{0.33}\\
    \bottomrule
\end{tabular}
}
\caption{Inter-annotator agreement for absolute-comparison AMT studies, using Krippendorff's $\alpha$ computed on an interval scale from -1 to 1. The aggregate score is computed by treating the annotations along each attribute as separate instances and computing agreement across the entire set.}
\label{table:agreement_absolute}
\end{table*}
\begin{table}[t!]
\centering
\small
\begin{tabular}{p{0.90\linewidth}}
\toprule
\textbf{SNLI \cite{bowman-etal-2015-large}} \\
\midrule[0.03em]
\textbf{Premise:} There are two kilted men, one of them older and is holding bagpipes with the other one with a drum.\\
\textbf{Hypothesis:} Two kiled (sp) men hold bagpipes\\
\textbf{Label:} contradiction \\
\textbf{NLL-Predicted Explanation:} The two kilted men are not holding bagpipes, they are holding a bagpipe and a drum.\\
\textbf{Filter-Predicted Explanation:} Just because there are kilted men does not necessarily mean that they are holding bagpipes. This can be seen from the older kilted man is holding the bagpipes rather than the kilder (sp) one.\\
\toprule
\textbf{CommonsenseQA \cite{talmor-etal-2019-commonsenseqa}} \\
\midrule[0.03em]
\textbf{Question:} The hardcovers were especially tall, so he removed a shelf on the what to make room? \\
\textbf{Answer choices:} hold alcohol, grocery store, \underline{bookcase}, nightstand, or chest of drawers \\
\textbf{NLL-Predicted Explanation:} Bookcases are tall, and so if they are full of hardcover books, they might need to be removed to make room for other things. \\
\textbf{Filter-Predicted Explanation:} People keep books on bookshelves, and bookcases are very commonly used to hold books. Since hardcover books can be pretty tall, it takes up extra vertical space which would require removal of an upper shelf of the bookcase. \\
\bottomrule
\end{tabular}
\caption{Randomly-selected instances that our filter model predicts differently than NLL at the ``select-1'' task, and got \textbf{incorrect} but NLL got correct. 
}
\label{tab:filter_incorrect_examples}
\end{table}
\begin{table*}
\resizebox{\textwidth}{!}{
\centering
\begin{tabular}{llrr}
    \toprule
    \textbf{AMT Study} & \textbf{Task/Dataset} & \textbf{\# Annotators} & \textbf{Median \# HITs (Mean)}\\
    \midrule
    GPT-3 Greedy w/ Dataset Prompts vs. Dataset & Com.QA/CoS-E & 16 & 31.5 (46.9) \\
    & Com.QA/ECQA & 13 & 35 (57.7)  \\
    & e-SNLI & 12 & 39 (62.5) \\
    \hline
    GPT-3 Greedy: Author-written vs. Dataset Prompts & Com.QA/CoS-E & 7 & 84 (107.1)  \\
    & Com.QA/ECQA & 13 & 49 (57.7)  \\
    & e-SNLI & 8 & 43.5 (93.8)  \\
    \hline
    GPT-3 Greedy w/ Author-written Prompts vs. Dataset & Com.QA/CoS-E & 8 & 90 (93.8)  \\
    & Com.QA/ECQA & 17 & 27 (44.1)  \\
    & e-SNLI & 8 & 93 (93.8)  \\
    \hline
    GPT-3 Greedy (Absolute) & Com.QA & 13 & 51 (57.7) \\
    & SNLI & 12 & 14 (62.5) \\
    \hline
    Dataset (Absolute) & CoS-E & 14 & 58 (53.6) \\
    & ECQA & 19 & 7 (39.5) \\
    & e-SNLI & 13 & 16 (57.7) \\
    \hline
    Acceptability (Training and Validation Data) & Com.QA (2973 HITs) & 34 & 70 (87.4) \\
    & SNLI (3000 HITs) & 14 & 128.5 (214.3) \\
    \hline
    Acceptability (Test Data) & Com.QA & 17 & 32 (44.1) \\
    & SNLI & 11 & 26 (68.1) \\
    & SNLI (Test2) & 7 & 65 (107.1) \\
    & CoS-E & 13 & 48 (57.7)\\
    & ECQA & 16 & 38.5 (46.9)\\
    & e-SNLI & 9 & 60 (83.3)\\
    \hline
    GPT-3 Filtered (Absolute) & Com.QA (468 HITs) & 10 & 44.5 (46.8) \\
    & SNLI (273 HITs) & 6 & 53 (45.5) \\
    \bottomrule
\end{tabular}
}
\caption{Total \# of annotators and mean \# HITs completed per-annotator for each AMT study (out of 750 total \# HITs unless otherwise specified = 3 annotators for each of 250 test instances).
}
\label{table:annotators}
\end{table*}
\begin{table*}
\centering
\resizebox{\linewidth}{!}{
\begin{tabular}{lccccccc}
    \toprule
    \textbf{Set of Test Explanations} & \textbf{Generality} & \textbf{Factuality} & \textbf{Grammar} & \textbf{New Info} & \textbf{Supp. Label} & \textbf{Amt. Info} & \textbf{Accept.} \\
    \midrule
    GPT-3 Greedy for Com.QA & $\mathbf{0.9_{0.4}}^\ddagger$ & $\mathbf{0.8_{0.4}}$ ($247$)$^\dagger$ & $\mathbf{1.0_{0.1}}^\ddagger$ & $0.1_{0.6}$ & $\mathbf{0.5_{0.7}}$ ($217$)$^\ddagger$ & $\mathbf{-0.1_{0.4}}$ ($186$)$^\ddagger$ & $\mathbf{0.5_{0.6}}\ddagger$\\
    CoS-E & $-0.2_{0.9}$ & $0.5_{0.5}$ ($131$) & $-0.3_{0.7}$ & $0.1_{0.8}$ & $-0.3_{0.9}$ ($190$) & $-0.5_{0.5}$ ($78$) & $-0.9_{0.4}$\\
    \hline
    GPT-3 Greedy for Com.QA & $\mathbf{0.9 _{0.4}}^\vee$ & $\mathbf{0.8_{0.4}}$ ($247$)$^\ddagger$ & $\mathbf{1.0_{0.1}}^\ddagger$ & $0.1_{0.6}$ & $0.5_{0.7}$ ($217$) & $-0.1_{0.4}$ ($186$) & $\mathbf{0.5_{0.6}}\ddagger$\\
    ECQA & $0.8_{0.4}$ & $0.6_{0.4}$ ($249$) & $0.1_{0.7}$ & $\mathbf{0.6_{0.5}}^\ddagger$ & $\mathbf{0.7_{0.5}}$ ($247$)$^\wedge$ & $\mathbf{0.5_{0.5}}$ ($239$)$^\ddagger$ & $0.1_{0.6}$\\
    \hline
    GPT-3 Greedy for SNLI & $\mathbf{0.7_{0.5}}^\wedge$ & $0.7_{0.5}$ ($246$) & $\mathbf{1.0_{0.2}}^\dagger$ & $0.1_{0.6}$ & $-0.1_{0.6}^*$ & $-0.2_{0.4}$ ($203$) & $0.0_{0.8}$\\
    e-SNLI & $0.6_{0.6}$ & $0.8_{0.4}$ ($236$) & $0.9_{0.4}$ & $\mathbf{0.2_{0.5}}^\vee$ & $\mathbf{0.2_{0.5}}^{*\ddagger}$ & $\mathbf{-0.1_{0.4}}$ ($238$)$^\wedge$ & $\mathbf{0.7_{0.4}}^\ddagger$\\
    \bottomrule
\end{tabular}
}
\caption{Statistics from the graphs plotted in \autoref{fig:absolute_all}. Mean $\pm$ standard error presented; numbers in parenthesis indicate the number of datapoints, if not 250. $^*$For SNLI, we modified the evaluation framework such that ``Supports Label'' was always answered instead of being conditioned on ``New Info''. Statistical significance results using a one-sided Wilcoxon signed-rank test at $p$-values of $\ddagger=0.00001$, $\dagger=0.0001$, $\vee=0.01$, $\wedge=0.1$ indicates that the median difference between the marked score distribution and the unmarked score distribution is greater than 0.
}
\label{table:absolute_stats}
\end{table*}
\begin{table*}
\centering
\resizebox{\textwidth}{!}{
\begin{tabular}{lccccccc}
    \toprule
    \textbf{Set of Test Explanations} & \textbf{Generality} & \textbf{Factuality} & \textbf{Grammar} & \textbf{New Info} & \textbf{Supp. Label} & \textbf{Amt. Info} & \textbf{Accept.} \\
    \midrule
    Com.QA Pred. \textbf{Correctly} & $\mathbf{0.9_{0.3}}$ ($127$)$^\vee$ & $\mathbf{0.9_{0.3}}$ ($126$)$^\vee$ & $1.0_{0.1}$ ($127$) & $0.0_{0.6}$ ($127$) & $\mathbf{0.7_{0.6}}$ ($108$)$^\dagger$ & $-0.1_{0.4}$ ($98$) & $0.5_{0.6}$ ($127$) \\
    Com.QA Pred. \textbf{Incorrectly} & $0.8_{0.4}$ ($123$) & $0.7_{0.5}$ ($121$) & $1.0_{0.1}$ ($123$) & $\mathbf{0.3_{0.6}}$ ($123$)$^\vee$ & $0.3_{0.8}$ ($109$) & $-0.1_{0.4}$ ($88$) & $0.5_{0.7}$ ($123$)  \\
    \hline
    SNLI Pred. \textbf{Correctly} & $\mathbf{0.8_{0.5}}$ ($115$)$^\wedge$ & $0.7_{0.5}$ ($112$) & $\mathbf{1.0_{0.1}}$ ($115$)$^\wedge$ & $-0.1_{0.7}$ ($115$) & $-0.2_{0.6}$ ($115$) & $-0.3_{0.4}$ ($83$) & $-0.1_{0.8}$ ($115$) \\
    SNLI Pred. \textbf{Incorrectly} & $0.7_{0.5}$ ($135$) & $0.7_{0.5}$ ($134$) & $1.0_{0.2}$ ($135$) & $\mathbf{0.2_{0.4}}$ ($135$)$^\ddagger$ & $\mathbf{0.1_{0.5}}$ ($135$)$^\vee$ & $\mathbf{-0.2_{0.4}}$ ($120$)$^\wedge$ & $\mathbf{0.1_{0.8}}$ ($135$)$^\wedge$ \\
    \bottomrule
\end{tabular}
}
\caption{Statistics from the graphs of GPT-3 greedy explanations plotted in \autoref{fig:absolute_pc_vs_pu}. See the caption of \autoref{table:absolute_stats} for further details.}
\label{table:pc_pu_stats}
\end{table*}
\begin{table*}
\centering
\resizebox{\linewidth}{!}{
\begin{tabular}{lccccccc}
    \toprule
    \textbf{Set of Test Explanations} & \textbf{Generality} & \textbf{Factuality} & \textbf{Grammar} & \textbf{New Info} & \textbf{Supp. Label} & \textbf{Amt. Info} & \textbf{Accept.} \\
    \midrule
    GPT-3 \textbf{Greedy} for Com.QA & $0.9_{0.4}$ ($156$) & $\mathbf{0.8_{0.4}}$ ($153$)$^\vee$ & $\mathbf{1.0_{0.1}}$ ($156$)$^\ddagger$ & $0.1_{0.6}$ ($156$) & $0.5_{0.7}$ ($135$) & $-0.1_{0.5}$ ($117$) & $0.3_{0.7}$ $(156)$\\
    GPT-3 \textbf{Filtered} for Com.QA & $\mathbf{0.9_{0.3}}$ ($156$)$^\wedge$ & $0.7_{0.3}$ ($155$) & $0.8_{0.4}$ ($156$) & $\mathbf{0.7_{0.4}}$ ($156$)$^\ddagger$ & $\mathbf{0.9_{0.4}}$ ($154$)$^\ddagger$ & $\mathbf{0.2_{0.3}}$ ($152$)$^\ddagger$ & $\mathbf{0.6_{0.6}}$ $(156)^\vee$\\
    \midrule
    GPT-3 \textbf{Greedy} for SNLI & $0.8_{0.4}$ ($91$) & $0.6_{0.6}$ ($91$) & $\mathbf{0.9_{0.3}}$ ($91$)$^\ddagger$ & $0.0_{0.6}$ ($91$) & $-0.2_{0.6}^*$ ($91$) & $-0.2_{0.5}$ ($66$) & $-0.5_{0.7}$ $(91)$ \\
    GPT-3 \textbf{Filtered} for SNLI & $0.8_{0.5}$ ($91$) & $0.7_{0.4}$ ($88$) & $0.7_{0.4}$ ($91$) & $\mathbf{0.5_{0.6}}$ ($91$)$^\ddagger$ & $\mathbf{0.5_{0.5}}^*$ ($91$)$^\ddagger$ & $\mathbf{0.0_{0.3}}$ ($89$)$^\dagger$ & $\mathbf{0.1_{0.8}}$  $(91)^\ddagger$\\
    \bottomrule
\end{tabular}
}
\caption{Statistics from the graphs plotted in \autoref{fig:absolute_greedy_vs_filtered}. See the caption of \autoref{table:absolute_stats} for further details.
}
\label{table:absolute_greedy_vs_filtered_stats}
\end{table*}
\begin{table*}[ht]
\centering
\small
\begin{tabular}{lrrrr}
    \toprule
     & \multicolumn{2}{c}{\textbf{``Select-1'' Acc@2/3}} & \multicolumn{2}{c}{\textbf{Explanation-level AP@2/3}}\\
     \cmidrule(lr){2-3}\cmidrule(lr){4-5}
     $\downarrow$\textbf{Model}/\textbf{Split}$\rightarrow$ & \textbf{Dev} & \textbf{Test} & \textbf{Dev} & \textbf{Test}\\
    \midrule
    Random & $ 57.3_{0.4}$ & $ 57.9_{0.4} $ & $56.2_{0.5}$ & $58.0_{0.9}$ \\
    Constant & --- & --- & $56.9$ & $58.0$ \\
    NLL & $79.1$ & $79.6$ & $ 77.5 $ & $ 75.0 $ \\
    \midrule[0.03em]
    T0-3B Expl.-only & $77.1_{3.5}$ & $75.8_{1.2}$ & $75.6_{2.0}$ & $77.3_{1.4}$ \\
    \midrule[0.03em]
    T0-3B & $\mathbf{86.6_{0.9}}$ & $\mathbf{85.8_{0.7}} $ & $\mathbf{85.6_{0.5}} $ & $\mathbf{87.0_{0.8}} $ \\
    \midrule
    Oracle Upper-Bound & $97.8$ & $97.6$ & $ 100.0 $ & $100.0$ \\
    \bottomrule
\end{tabular}
\caption{Results for acceptability classifiers trained on CommonsenseQA, with ``acceptability" defined as: ``at least 2/3 annotators labelled as acceptable." 
Subscripts indicate standard error over models trained with 5 different random seeds.
The oracle upper bound is based on dataset properties (\sect{ssec:room_for_improvement}).
}
\label{table:accept_results_commonsenseqa_2outof3}
\end{table*}
\begin{table*}[ht]
\centering
\small
\begin{tabular}{lrrrrrr}
    \toprule
    & \multicolumn{3}{c}{\textbf{``Select-1'' Acc@2/3}} & \multicolumn{3}{c}{\textbf{Explanation-level AP@2/3}}\\
    \cmidrule(lr){2-4}\cmidrule(lr){5-7}
    $\downarrow$\textbf{Model}/\textbf{Split}$\rightarrow$ & \textbf{Dev} & \textbf{Test} & \textbf{Test2} & \textbf{Dev} & \textbf{Test} & \textbf{Test2} \\
    \midrule
    Random & $28.2_{0.5}$ & $27.8_{0.2}$ & $28.0_{0.1}$ & $28.1_{0.9}$ & $27.6_{0.3}$ & $28.3_{0.6}$ \\
    Constant & --- & --- & --- & $28.2$ & $27.8$ & $28.0$\\
    NLL & $51.0$ & $51.2$ & $50.4$  & $47.7$ & $47.5$ & $46.1$\\
    \midrule[0.03em]
    T0-3B Expl.-only & $47.0_{1.0}$ & $50.5_{2.1}$ & $50.6_{2.8}$  & $48.9_{1.4}$ &$45.2_{1.5}$ & $44.9_{2.1}$\\
    \midrule[0.03em]
    T0-3B & $\mathbf{57.8_{1.9}}$ & $\mathbf{60.3_{1.5}}$ & $\mathbf{59.2_{2.3}}$ & $\mathbf{66.7_{3.3}}$ & $\mathbf{64.7_{3.3}}$ &  $\mathbf{67.1_{3.6}}$ \\
    \midrule
    Oracle Upper-Bound & $76.0$ & $81.2$ & $77.6$ & $100.0$ & $100.0$ & $100.0$ \\
    \bottomrule
\end{tabular}
\caption{Results for acceptability classifiers trained on SNLI with ``acceptability" defined as: ``at least 2/3 annotators labelled as acceptable." Subscripts indicate standard error over models trained with 5 different random seeds. The oracle upper bound is based on dataset properties (\sect{ssec:room_for_improvement}).}
\label{table:accept_results_snli_2outof3}
\end{table*}

\section{Additional Filter Results}
\label{sec:app:additional_filter_results_2outof3}

In the main experiments, at evaluation time, we labelled an explanation as acceptable if $3/3$ annotators agreed on it. Here, we report results if this threshold is relaxed to $2/3$. Overall, the results are comparable: T0-3B outperforms the baselines according to both select-1 accuracy and AP (see \autoref{table:accept_results_commonsenseqa_2outof3} and \autoref{table:accept_results_snli_2outof3}). 

\clearpage

\end{document}